\pdfoutput=1

\documentclass[letterpaper, 10 pt, journal, twoside]{IEEEtran}
\usepackage[T1]{fontenc}
\usepackage{cite}
\usepackage{amssymb,amsfonts}
\usepackage{blindtext}
\usepackage{hyperref}
\usepackage{algorithmic}
\usepackage{graphicx}
\usepackage{textcomp}
\usepackage{mathtools}
\usepackage{changes}
\usepackage{subcaption}
\usepackage{algorithm}
\usepackage{xcolor}
\usepackage{color, soul}
\usepackage{amsmath}
\usepackage{booktabs}
\usepackage{multirow}
\usepackage{xstring}
\usepackage{hhline}
\usepackage{kotex}
\usepackage{siunitx}
\usepackage{comment}
\usepackage{physics}
\usepackage{tikz}
\usepackage{cleveref}
\crefformat{section}{\S#2#1#3} 
\crefformat{subsection}{\S#2#1#3}
\crefformat{subsubsection}{\S#2#1#3}

\floatname{algorithm}{Procedure}

\definecolor{qwe}{RGB}{0, 0, 255}
\makeatletter
\DeclareRobustCommand{\iscircle}{\mathord{\mathpalette\is@circle\relax}}
\newcommand\is@circle[2]{%
  \begingroup
  \sbox\z@{\raisebox{\depth}{$\m@th#1\bigcirc$}}%
  \sbox\tw@{$#1\square$}%
  \resizebox{!}{\ht\tw@}{\usebox{\z@}}%
  \endgroup
}
\makeatother

\IEEEoverridecommandlockouts                              

\usepackage{cite}
\usepackage{caption}
\newcommand{\rom}[1]{\uppercase\expandafter{\romannumeral #1\relax}}

\begin{document}

\title{Patchwork: Concentric Zone-based Region-wise Ground Segmentation with Ground Likelihood Estimation Using a 3D LiDAR Sensor}

\author{Hyungtae Lim$^{1}$, \textit{Student Member, IEEE}, Minho Oh$^{1}$, Hyun Myung$^{1}$, \textit{Senior Member, IEEE}
\thanks{This work was supported by the Industry Core Technology Development Project, 20005062, Development of Artificial Intelligence Robot Autonomous Navigation Technology for Agile Movement in Crowded Space, funded by the Ministry of Trade, Industry \& Energy (MOTIE, Republic of Korea) and by the research project “Development of A.I. based recognition, judgement and control solution for autonomous vehicle corresponding to atypical driving environment,” which is financed from the Ministry of Science and ICT (Republic of Korea) Contract No. 2019-0-00399. The students are supported by the BK21 FOUR from the Ministry of Education (Republic of Korea).} 
\thanks{$^{1}$Hyungtae Lim, $^{1}$Minho Oh, and $^{1}$Hyun Myung are with the School of Electrical Engineering, KI-AI, KI-R at KAIST (Korea Advanced Institute of Science and Technology), Daejeon, 34141, South Korea. {\tt\small \{shapelim, minho.oh, hmyung\}@kaist.ac.kr} \hfill \break 
}%
}

\captionsetup[figure]{labelformat={default},labelsep=period,name={Fig.}}

\markboth{ }
{Lim \MakeLowercase{\textit{et al.}}: Patchwork: Concentric Zone-based Region-wise Ground Segmentation with Ground Likelihood Estimation} 
\maketitle
\IEEEpeerreviewmaketitle


\begin{abstract}

Ground segmentation is crucial for terrestrial mobile platforms to perform navigation or neighboring object recognition. Unfortunately, the ground is not flat, as it features steep slopes; bumpy roads; or objects, such as curbs, flower beds, and so forth. To tackle the problem, this paper presents a novel ground segmentation method called \textit{Patchwork}, which is robust  for addressing the under-segmentation problem and operates at more than 40 Hz. In this paper, a point cloud is encoded into a Concentric Zone Model--based representation to assign an appropriate density of cloud points among bins in a way that is not computationally complex. This is followed by Region-wise Ground Plane Fitting, which is performed to estimate the partial ground for each bin. Finally, Ground Likelihood Estimation is introduced to dramatically reduce false positives. As experimentally verified on SemanticKITTI and rough terrain datasets, our proposed method yields promising performance compared with the state-of-the-art methods, showing faster speed compared with existing plane fitting--based methods. Code is available: \href{https://github.com/LimHyungTae/patchwork}{\texttt{https://github.com/LimHyungTae/patchwork}}
\end{abstract}


\vspace{-0.5cm}
\section{Introduction} \label{sec:intro}
\vspace{-0.1cm}

\IEEEPARstart{I}{n} recent years, \textcolor{black}{there has been an increased demand} to perceive surroundings for mobile platforms, such as Unmanned Ground Vehicles (UGVs), Unmanned Aerial Vehicles (UAVs), or autonomous cars\textcolor{black}{. To accomplish this,} numerous researchers have applied various 3D perception methods \cite{behley2019semantickitti, lim2020normal,byun2015drivable, douillard2011segmentation}. In particular, a 3D \textcolor{black}{light detection and ranging (LiDAR)} sensor has been extensively deployed \textcolor{black}{due to allowing for} centimeter-level accuracy and omnidirectional sensing, as well as \textcolor{black}{its} ability to measure great distances compared with stereo cameras \cite{behley2019semantickitti, thrun2006stanley, moosmann2009segmentation}. Accordingly, a 3D point cloud captured by a LiDAR sensor is utilized \textcolor{black}{for} semantic segmentation \cite{milioto2019rangenet++, zermas2017fast}, tracking \cite{asvadi2015detection_roof}, detection  \cite{ali2018yolo3d}, and so forth.

In this paper, we \textcolor{black}{specifically} focus on ground segmentation tasks \cite{himmelsbach2010fast, narksri2018slope}. There are two main purposes of ground segmentation. One is to estimate the movable area \cite{na2016drivable, byun2015drivable} for successful navigation. The other purpose, {on which this paper places more emphasis,} is the segmentation of a point cloud to recognize or track moving objects. Terrestrial vehicles or humans inevitably come into contact with the ground \cite{lim21erasor}; ideally, dynamic objects can be recognized in a simple way, such as \textcolor{black}{through} Euclidean clustering if the ground is well estimated \cite{zermas2017fast, asvadi20163d}. Furthermore, because most cloud points belong to the ground, ground segmentation can significantly reduce computational power when \textcolor{black}{one is} performing object segmentation or detection in a preprocessing stage \cite{cheng2020simple}. Thus, \textit{ground} in this study \textcolor{black}{refers to} not only the road, which is a movable area, but also all regions \textcolor{black}{that} moving objects can come into contact with, including sidewalks or lawns. 

\begin{figure}[t!]
    \centering
	\begin{subfigure}[b]{0.5\textwidth}
		\includegraphics[width=1.0\textwidth]{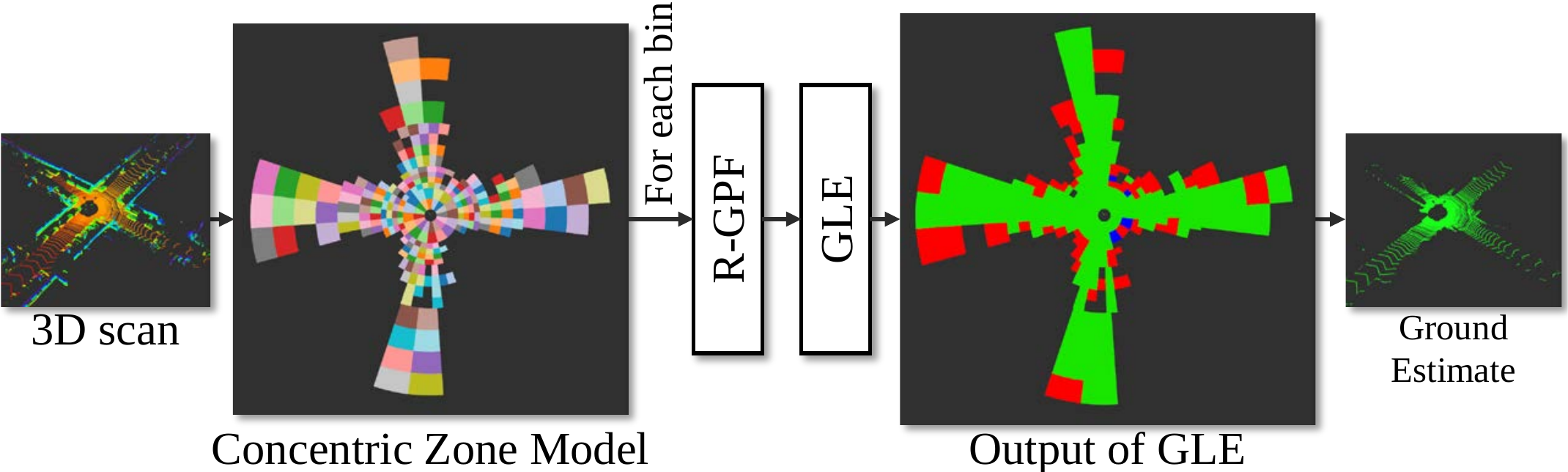}
	\end{subfigure}
	\vspace{-0.6cm}
	\caption{Overview of our proposed method called \textit{Patchwork}. Patchwork mainly consists of three parts: Concentric Zone Model \textcolor{black}{(CZM)}--based polar grid representation, Region-wise Ground Plane Fitting (R-GPF), and Ground Likelihood Estimation (GLE).}
	\label{fig:overview}
	\vspace{-0.4cm}
\end{figure}


In this study, as presented in Fig.~\ref{fig:overview}, we propose a novel Concentric Zone Model \textcolor{black}{(CZM)}--based region-wise ground segmentation method, called \textit{Patchwork}, which is an extension of Region-wise Ground Plane Fitting (R-GPF) in our previous study \cite{lim21erasor}. \textcolor{black}{T}he aim of R-GPF \textcolor{black}{in our previous study} was to estimate the ground points for static map building purposes, whereas here, we focus only on ground segmentation on a 3D point cloud. We also conduct detailed experiments on the impact of the \textcolor{black}{bin} size, which was not covered in our previous paper.

In summary, the contribution of this paper is threefold: 
\begin{itemize}
	\item To the best of our knowledge, it is the first attempt to analyze the impact of \textcolor{black}{bin} size when estimating ground planes in complex urban environments using the SemanticKITTI dataset \cite{behley2019semantickitti}. Accordingly, an efficient\textcolor{black}{,} non-uniform\textcolor{black}{,} region-wise representation \textcolor{black}{of} a 3D point cloud is proposed, referred to as \textcolor{black}{a} \textcolor{black}{CZM}--based representation whose bin size is different depending on each zone.
	\item Also, we leverage Ground Likelihood Estimation (GLE) in terms of \textit{uprightness}, \textit{\textcolor{black}{elevation}}, and \textit{flatness} to determine whether each bin is ground.
	\item Our proposed method shows promising performance over the state-of-the-art\textcolor{black}{,} region-wise fitting--based methods \textcolor{black}{at more than} 40 Hz. In particular, Patchwork estimates the ground points with the least \textcolor{black}{recall} variance, which \textcolor{black}{shows} that our proposed method overcomes the under-segmentation issue in complex urban environments.
\end{itemize}


\section{Related Works}
\color{black}
\vspace{-0.125cm}
\subsection{The Difficulties of Ground Segmentation}
\vspace{-0.125cm}

One may argue that it is a simple task that can be easily estimated by filtering a point cloud based on sensor height or using RANSAC \cite{fischler1981ransac} which is a renowned method \textcolor{black}{for} estimat\textcolor{black}{ing a} plane. Unfortunately, there are three main issues that impede algorithms from conducting precise ground segmentation: a) \textcolor{black}{t}here exists a partially steep slope or bumpy road, b) curbs or flower beds make some regions uneven, and c) because all surrounding objects are taken into account as outliers in the ground segmentation tasks, these objects hinder plane fitting. For these reasons, sometimes under-segmentation occurs, in which \textcolor{black}{case} points belonging to different objects are merged into the same segment \cite{narksri2018slope, moosmann2009segmentation}.

\vspace{-0.125cm}
\subsection{Ground Plane Estimation Methods}
\vspace{-0.125cm}

To tackle these issues, numerous researchers have studied various approaches. For instance, Douillard \textit{et al.} \cite{douillard2011segmentation} and Chen \textit{et al.} \cite{chen2014gaussian} employed Gaussian process--based methods. On the other hand, Tse \textit{et al.} \cite{tse2012unified}, Byun \textit{et al.} \cite{byun2015drivable}, and Rummelhard \textit{et al.} \cite{rummelhard2017ground} proposed Markov Random Field--based methods. These methods can \textcolor{black}{be used to} estimate detailed ground points yet requiring much computational time, so it may not be \textcolor{black}{appropriate} to use them as preprocessing algorithms whose speed should be guaranteed at \textcolor{black}{more than} 20 Hz.

\vspace{-0.125cm}
\subsection{Scan Representation}
\vspace{-0.125cm}

Meanwhile, grid representation--based methods have been widely utilized to leverage expressibility compared with singular plane model--based methods \cite{douillard2011segmentation, asvadi2015detection_roof}. In particular, polar grid representation\textcolor{black}{,} which treats a point cloud in cylindrical coordinates\textcolor{black}{,} is commonly employed these days because it naturally compensates for the geometric characteristics of 3D LiDAR sensors \cite{himmelsbach2010fast, steinhauser2008motion, lim21erasor, narksri2018slope, cheng2020simple}. In practice, Thrun \textit{et al.} \cite{thrun2006stanley} presented a grid cell--based binary ground classification method in a probabilistic way to predict the movable area for autonomous driving in the DARPA challenge. These methods are mainly divided into two categories: a) elevation map--based and b) model fitting--based methods. Accordingly, the latter category can be further classified into two main methodologies: a) line fitting--based and b) plane fitting--based methods.

\vspace{-0.125cm}
\subsection{Elevation Map--based 2.5D Grid Representation}
\vspace{-0.125cm}

First, elevation map--based methods \textcolor{black}{are used to} distinguish between ground and non-ground points by encoding a 3D point cloud into 2.5D grid representations \cite{asvadi2015detection_roof,thrun2006stanley}. Thrun \textit{et al.} \cite{thrun2006stanley} utilized relative height and Asvadi \textit{et al.} \cite{asvadi2015detection_roof} used average height and its covariance on each grid. These methods have strong advantages over other methods in terms of speed and computational cost. However, there are some potential risks that {sometimes a steep slope region could be considered as a non-ground region because of large {$z$} value difference between its supremum and infimum points with respect to Z-axis.}

\vspace{-0.125cm}
\subsection{Multiple Line Fitting--based Ground Segmentation}
\vspace{-0.125cm}

Next, Himmelsbach \textit{et al.} \cite{himmelsbach2010fast} and Steinhauser \textit{et al.} \cite{steinhauser2008motion} introduced 2D line fitting on \textcolor{black}{a} uniform polar grid representation to estimate the straight-line equation on each grid. Then, in each grid, it \textcolor{black}{was} determined whether points \textcolor{black}{were} ground points \textcolor{black}{by comparing} between constant thresholds and the estimated parameters, such as \textcolor{black}{the} point-to-line distance, gradient, or y-intercept. 

\subsection{Multiple Plane Fitting--based Ground Segmentation}
\vspace{-0.125cm}

\textcolor{black}{Sharing} their view\textcolor{black}{s} of region-wise fitting yet improving robustness, other researchers have conducted region-wise plane fitting--based approaches \cite{zermas2017fast, narksri2018slope, cheng2020simple, lim21erasor}. For instance, Zermas \textit{et al.} \cite{zermas2017fast} divided a point cloud into three parts along the x-axis of the body frame, which is \textcolor{black}{the} forward direction of a vehicle. This method is based on the premise that a slope usually changes along the x-axis; however, this assumption sometimes fails when it comes to a bumpy road or a complex intersection. To resolve the problem, Narksri \textit{et al.} \cite{narksri2018slope} proposed a slope-robust method using consecutive ring patterns in the scan data as well as the concept of \textcolor{black}{the} continuity of the region-wise estimated plane along the radial direction. Furthermore, Narksri \textit{et al.} \cite{narksri2018slope} and Cheng \textit{et al.} \cite{cheng2020simple} proposed an adaptive way \textcolor{black}{of setting} a grid size depending on the density of the cloud points or \textcolor{black}{the} incidence angle.

\subsection{Deep Learning-based Methods}
\vspace{-0.125cm}

Of course, as the deep learning era has come, Milioto \textit{et al.} \cite{milioto2019rangenet++} proposed RangeNet++ to estimate point-wise labels on a 3D point cloud and Paigwar \textit{et al.} \cite{paigwar2020gndnet} presented GndNet\textcolor{black}{,} which estimates ground plane elevation information in a grid--based representation to discern ground points in real time. Unfortunately, {these methods usually require high computational resource{s}}. In addition, these methods tend to be highly fitted to the environments of train dataset; thus, {the performance of those can be potentially degraded when used in quite different environments from the training {dataset} or different sensor configuration \cite{wong2020identifying}.}

\section{Methodology of Patchwork}
\vspace{-0.1cm}
 
The following paragraphs highlight the problem definition and the reasoning behind each module of Patchwork. Patchwork mainly consists of three parts: CZM, \textcolor{black}{R-GPF}, and \textcolor{black}{GLE}.

\subsection{Problem Definition}

First, we begin by denoting a point cloud at the moment as  $\mathcal{P}$. Then, let $\mathcal{P}=\{\mathbf{p}_{1}, \mathbf{p}_{2}, \dots, \mathbf{p}_{k}, \dots, \mathbf{p}_{N}\}$ be a set of cloud points that contain $N$ points \textcolor{black}{at} the moment acquired by a 3D LiDAR sensor, where each point $\mathbf{p}_k$ consists of $\mathbf{p}_{k}=\{x_k, y_k, z_k\}$ in the Cartesian coordinates. In this paper, $\mathcal{P}$ is definitely classified into two classes: a set of ground points, $G$, and its complement, $G^{c}$, which satisfy $G \cup {G^c} = \mathcal{P}$. Note that ${G^c}$ denotes non-ground points, including vehicles, walls, street trees, pedestrians, and so forth.

Next, estimated ground points could be defined as $\hat{G}$. Because the estimation unavoidably contains inherent errors \cite{lim2020normal}, some points that are, in fact, from non-ground objects could be included in $\hat{G}$, and vice versa. 
In summary, $\hat{G}$ and $\hat{G}^c$ are expressed as follows:

\begin{equation}
\hat{G}=\text{TP}\cup\text{FP} \: \; \text{and} \: \; \hat{G}^c=\text{FN} \cup \text{TN}
\end{equation} 
where $\hat{G}$ and $\hat{G}^c$ also satisfy $\hat{G} \cup {\hat{G}^c} = \mathcal{P}$, and TP, FP, FN, and TN denote set\textcolor{black}{s} of \textit{true positives}, \textit{false positives}, \textit{false negatives}, and \textit{true negatives}, respectively. Thus, our goal is to discern $\hat{G}$ and $\hat{G}^c$ from a point cloud $\mathcal{P}$ \textcolor{black}{while} estimating as few FP\textcolor{black}{s} and FN\textcolor{black}{s} as possible.

\subsection{Concentric Zone Model}

As mentioned earlier, most multiple plane--based methods are based on the general assumption that the observable world might not be flat. Therefore, ground plane estimation should be conducted region-wise by assuming that small parts, or bins, of the possibly non-flat world and the ground can indeed be flat within the region.

\begin{figure}[t!]
	\centering 
	\begin{subfigure}[b]{0.15\textwidth}
		\includegraphics[width=1.0\textwidth]{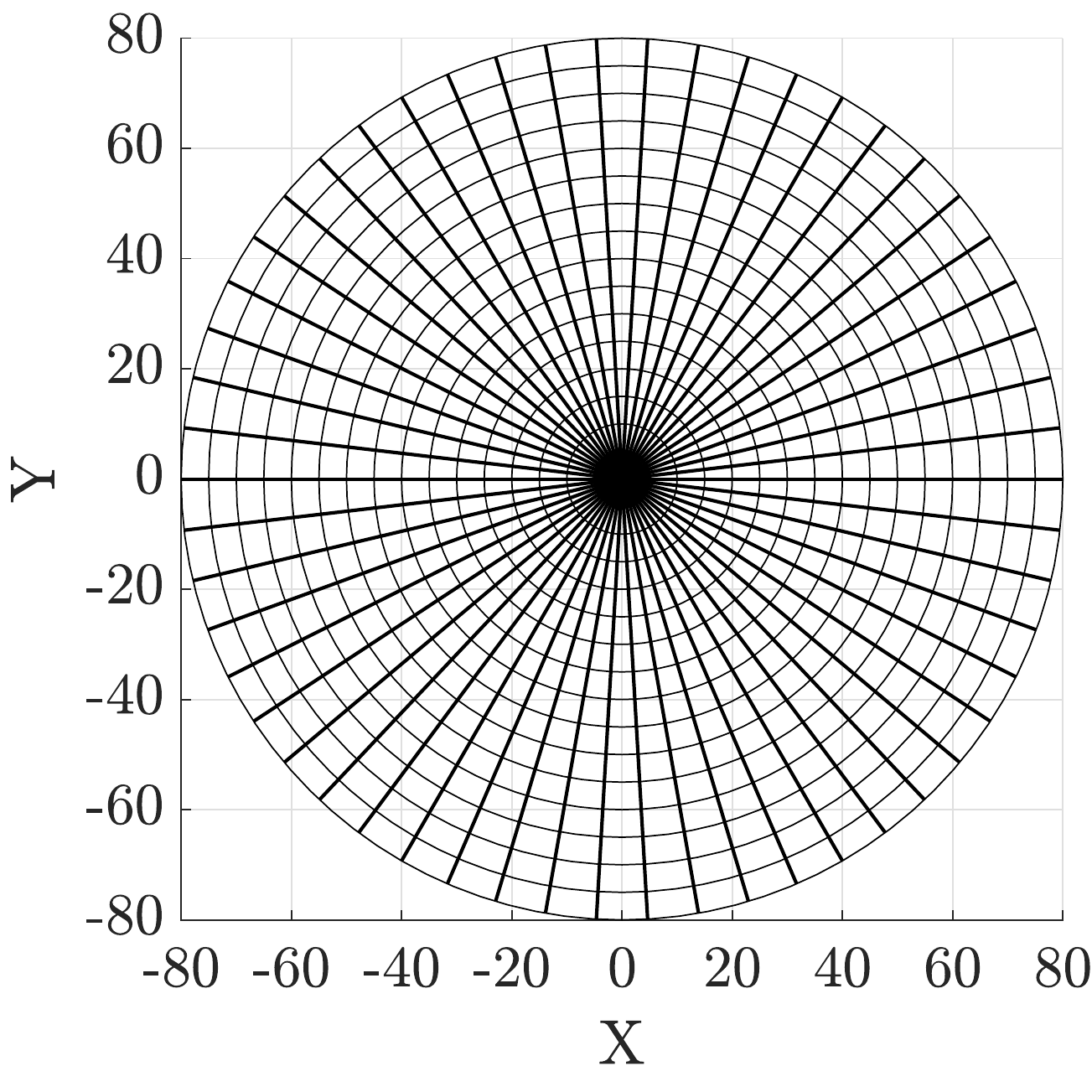}
		\caption{}
	\end{subfigure}
	\begin{subfigure}[b]{0.15\textwidth}
		\includegraphics[width=1.0\textwidth]{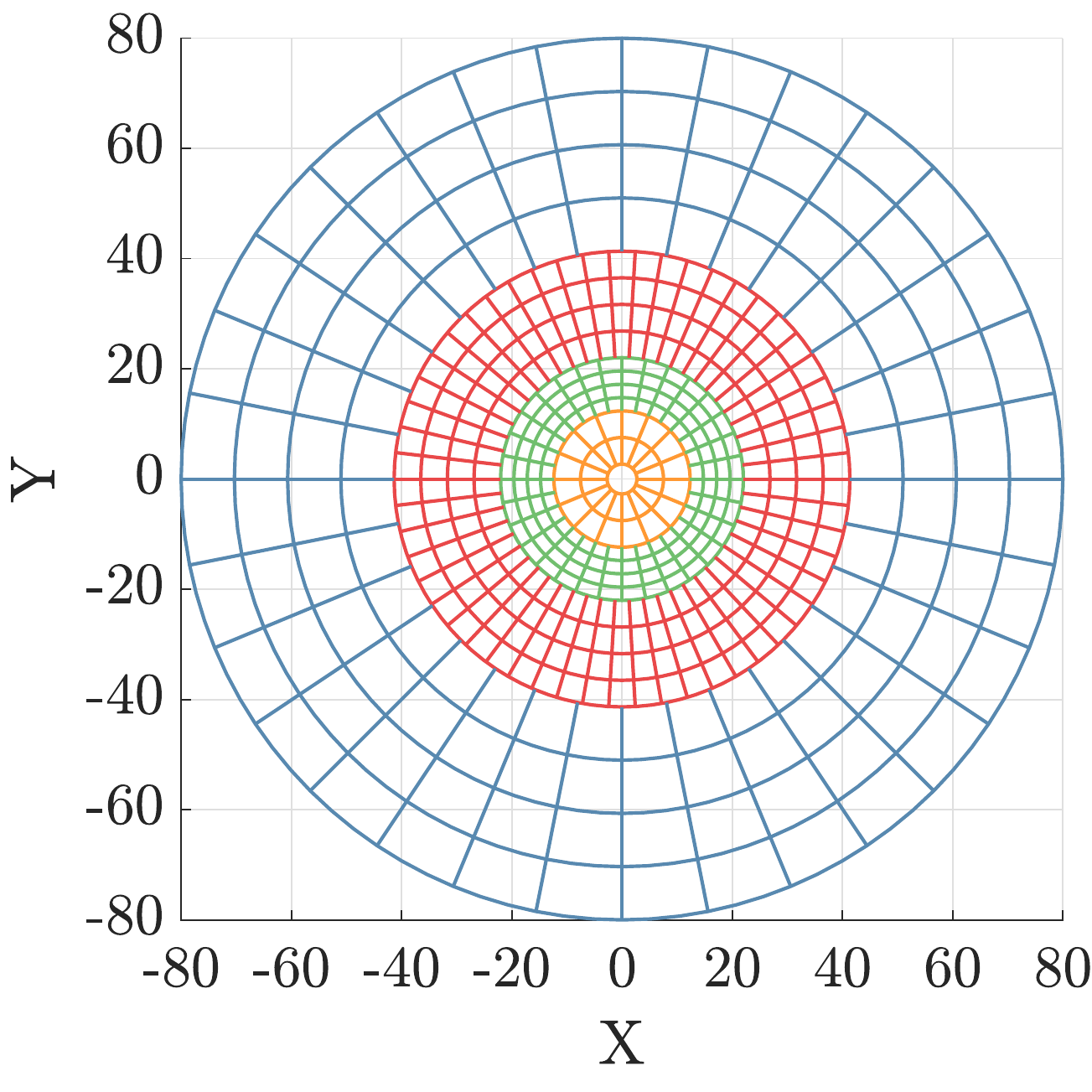}
		\caption{}
	\end{subfigure}
	\begin{subfigure}[b]{0.145\textwidth}
		\includegraphics[width=1.0\textwidth]{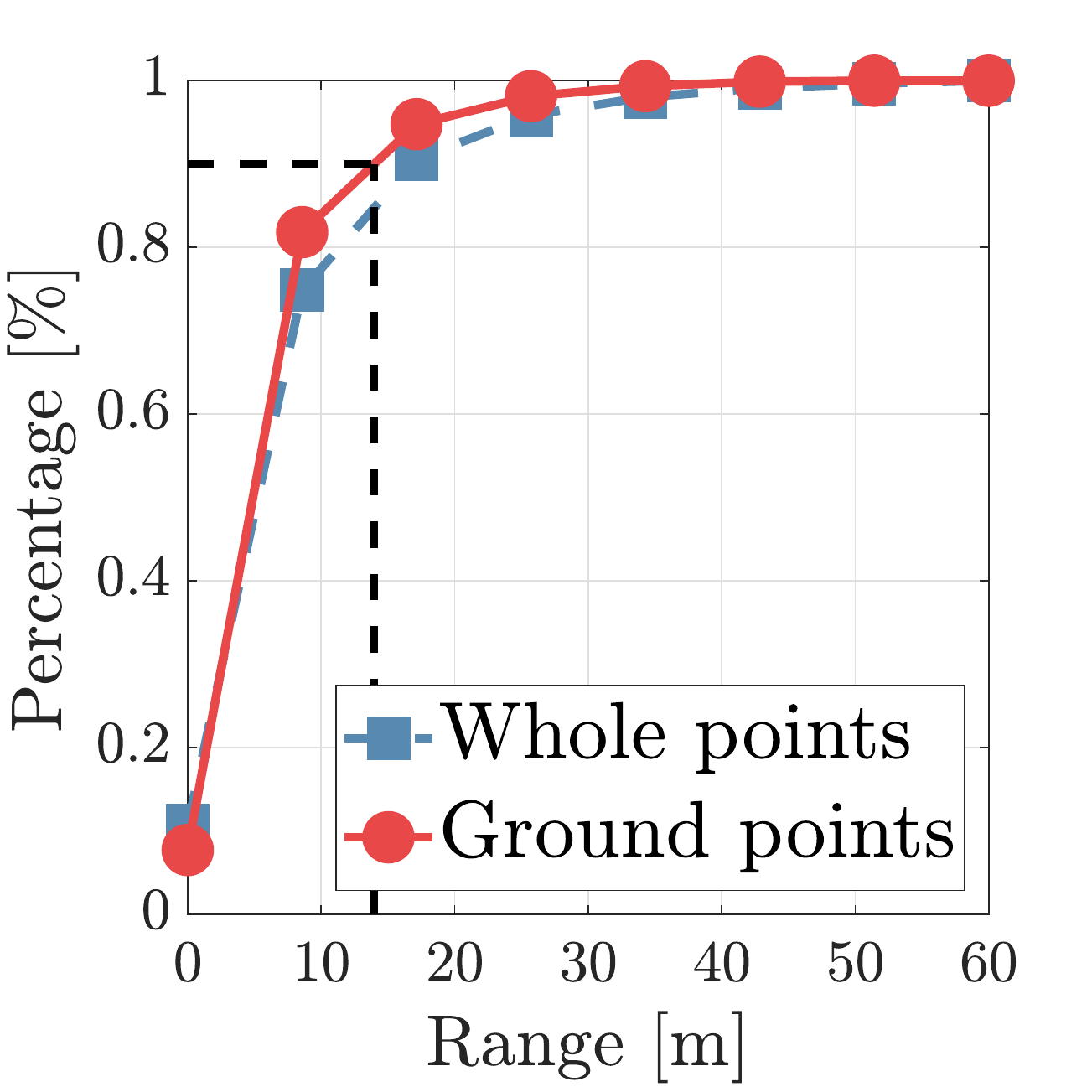}
		\caption{}
	\end{subfigure}
	\vspace{-0.20cm}
	\caption{(a) The uniform polar grid description \cite{himmelsbach2010fast,steinhauser2008motion,lim21erasor} (b) Our \textcolor{black}{CZM}--based polar grid description (c) Cumulative Distribution Function (CDF) of the range, in which \textcolor{black}{more than} 90\% of ground points are located within 20 m.}
	\label{fig:concentric_zone_model}
\end{figure}

Accordingly, some previous approaches utilized uniform polar grid representation, or $S$, to divide a point cloud into multiple bins with regular intervals of radial and azimuthal directions, i.e. \textit{rings} and \textit{sectors} \cite{lim21erasor}. More concretely, let $N_{r}$ and $N_{\theta}$ be the numbers of rings and sectors, respectively. Then, $S$ is divided into equal-sized bins whose size is $L_{\max}/N_{r}$ in the radial direction, where $L_{\max}$ denotes the maximum boundary, and  $2 \pi /N_{\theta}$ in the azimuthal direction as shown in Fig.~\ref{fig:concentric_zone_model}(a).


Unfortunately, as shown in Fig.~\ref{fig:concentric_zone_model}(c), the experimental evidence, which is measured on whole sequences on the SemanticKITTI dataset \cite{behley2019semantickitti} to take generalization into account, shows that most ground points are located close to the sensor frame. That is, \textcolor{black}{more than} 90\% of points belonging to the ground are located within 20 m.

For this reason, $S$ has two limitations. First, as the distance \textcolor{black}{becomes} farther away, a point cloud becomes too sparse to find the right ground plane, which we refer to as the \textit{sparsity issue}. \textcolor{black}{S}ome approaches adaptively adjust the size of a bin to cope with the logarithmic \textcolor{black}{point} distribution \cite{narksri2018slope,cheng2020simple,himmelsbach2010fast}. \textcolor{black}{However}, the \textcolor{black}{bin} size increases in a linear or quadratic way, so the sparsity issue is not completely resolved. On the other hand, when the size of \textcolor{black}{the} bins located \textcolor{black}{close} to the origin is too small to represent a unit space in $S$, \textcolor{black}{this} sometimes leads to \textcolor{black}{the} failure of \textcolor{black}{the} right normal vector estimation of the ground plane, which we refer to as the \textit{representability issue}.

\textcolor{black}{For} address\textcolor{black}{ing} these issues, the \textcolor{black}{CZM}--based polar grid representation, which is denoted as $\mathcal{C}$, is proposed to assign \textcolor{black}{the} appropriate density among bins \textcolor{black}{in a way that is not} computationally complex. \textcolor{black}{Accordingly}, $\mathcal{P}$ is divided into multiple zones, each of which is composed of bins of different sizes, as shown in Fig.~\ref{fig:concentric_zone_model}(b). Let $\left< N \right>=\{1,2,\dots,N\}$, then our proposed model is defined as follows:

\vspace{-0.30cm}
\begin{equation}
\mathcal{C}=\bigcup_{m\in\left<N_{Z}\right>} Z_{m}
\vspace{-0.1cm}
\end{equation}
where $Z_{m}$ denotes the $m$-th zone of $\mathcal{C}$ and $N_{Z}$ denotes the number of zones, which is empirically set to \textcolor{black}{four} in this paper. Let $Z_{m}=\left\{\mathbf{p}_{k} \in \mathcal{P} \mid L_{\min, m} \leq \rho_{k}<L_{\max, m}\right\}$\textcolor{black}{,} where $L_{\min, m}$ and $L_{\max, m}$ denote the minimum and \textcolor{black}{the} maximum radial boundary of $Z_{m}$, respectively; then, $Z_{m}$ is also divided into $N_{r, m}\times N_{\theta, m}$ bins, where each zone has a different bin size. Accordingly, each bin $\mathcal{S}_{i,j,m}$ is defined as follows:

\vspace{-0.25cm}
\footnotesize

\begin{equation}
\begin{aligned}
\mathcal{S}_{i,j,m}= \left\{ \mathbf{p}_{k} \in Z_{m} \mid \frac{(i-1) \cdot \Delta L_{m}}{N_{r, m}} \leq \rho_{k}-L_{\min, m}<\frac{i \cdot \Delta L_{m}}{N_{r, m}},\right.\\
\left.\frac{(j-1) \cdot 2 \pi}{N_{\theta, m}}-\pi \leq \theta_{k}<\frac{j \cdot 2 \pi}{N_{\theta, m}}-\pi \right\}.
\end{aligned} 
\vspace{-0.1cm}
\end{equation}

\normalsize

\noindent \textcolor{black}{where} $\rho_{k}=\sqrt{x_{k}^{2}+y_{k}^{2}}$, $\theta_k=\arctan2(y_k, x_k)$, $\Delta L_{m}=L_{\max, m}-L_{\min, m}$, and $L_{\max, m}=L_{\min, m+1}$ on $m=1,2,3$. Note that $L_{\max, 4}=L_{\max}$ and $L_{\min, 1}=L_{\min}$, for which the global minimum boundary $L_{\min}$ is \textcolor{black}{used} to consider emptiness over the vicinity of mobile platforms or vehicles. In fact, $Z_1$, $Z_2$, $Z_3$, and $Z_4$ are called the \textit{central zone}, \textit{quater zone}, \textit{half zone}, and \textit{outer zone}, respectively. Accordingly, $L_{\min, 2}=\frac{7 \cdot L_{\min} + L_{\max}}{8}$, $L_{\min, 3}=\frac{3 \cdot L_{\min} + L_{\max}}{4}$, and $L_{\min, 4}=\frac{L_{\min} + L_{\max}}{2}$. 

Note that the sizes of bins in $Z_1$ and $Z_4$ are set larger to resolve the sparsity issue and representability issue. Accordingly, $\mathcal{C}$ improves expressibility and thus allows robust estimation of a normal vector compared with the existing uniform representations, \textcolor{black}{thus} preventing under-segmentation \cite{lim21erasor, himmelsbach2010fast}. Furthermore, it reduces the actual number of bins, for instance, from 3,240 in $S$ to 504 in $\mathcal{C}$, thus enabling operation at \textcolor{black}{more than} 40 Hz (see Section \rom{4}.\textit{E}).

\subsection{Region-wise Ground Plane Fitting}

 Thereafter, each bin assigns \textcolor{black}{an} estimated partial ground via R-GPF; then, the partial ground points are merged later. In this paper, Principal Component Analysis (PCA) is utilized rather than using RANSAC. Of course, RANSAC tends to be less sensitive to outliers compared with PCA \cite{nurunnabi2014diagnostics,fischler1981ransac}. However, using PCA shows much faster speed than using RANSAC and shows acceptable performance \cite{zermas2017fast}; thus, PCA--based estimation is more appropriate as a preprocessing process. Additionally, experiments show that PCA--based methods are at least two times faster than RANSAC--based methods (see Section \rom{4}.\textit{E}).
 
 Given a bin, let $C \in \mathbb{R}^{3 \times 3}$ be the covariance matrix of cloud points in the unit space; three eigenvalues $\lambda_\alpha$ and the corresponding three eigenvectors ${\mathbf{v}}_\alpha$ are calculated as follows:

\vspace{-0.4cm}
\begin{equation}
    C \; {\mathbf{v}}_\alpha=\lambda_\alpha \; {\mathbf{v}}_\alpha \label{eq:eigen_decomposition}
    \vspace{-0.1cm}
\end{equation}
\text{where} $\alpha=1,2,3$, and assume $\lambda_1 \geq \lambda_2 \geq \lambda_3$. Then, the eigenvector that corresponds to the smallest eigenvalue, i.e. ${\mathbf{v}}_3$, is the most likely to represent the normal vector to the ground plane. 
Therefore, let $\mathbf{n}={\mathbf{v}}_3 = [a \ b \ c]^T$; then, the plane coefficient can be calculated as $d=-\mathbf{n}^{T} \bar{\mathbf{p}}$, where $\bar{\mathbf{p}}$ denotes the mean point of a unit space. 

 
 For simplicity, let the $n$-th bin be $S_{n}$ over all bins\textcolor{black}{,} whose number is equal to $N_{\mathcal{C}}=\sum_{m=1}^{N_{Z}}N_{r,m}\times N_{\theta,m}$. If the cardinality of $S_{n}$ is large enough, the lowest-height points are selected to be used as initial seeds. Indeed, the points of each bin with the lowest heights are most likely to belong to the ground surface \cite{himmelsbach2010fast, zermas2017fast, narksri2018slope}. Let $\bar{z}_{\text{init}}$ be the mean {$z$} value of the total $N_{\text{seed}}$ number of the selected seed point; then, initial estimated ground points set $\hat{G}^{0}_{n}$ is obtained as follows:

\vspace{-0.45cm}
\begin{equation}
\begin{aligned}
\hat{G}^{0}_{n}=\left\{ \mathbf{p}_{k} \in S_{n} \mid z(\mathbf{p}_{k}) < \bar{z}_{\text{init}} + z_{\text{seed}} \right.\}
\end{aligned}
\vspace{-0.15cm}
\end{equation}
where $z(\cdot)$ returns the {$z$} value of a point and $z_{\text{seed}}$ denotes the height margin. 

Because our method is iterative, let \textcolor{black}{the} estimated ground points set on the $l$-th iteration be $\hat{G}^{l}_{n}$. Then, the normal vector of $\hat{G}^{l}_{n}$, which is denoted as $\mathbf{n}^{l}_{n}$, is obtained through (\ref{eq:eigen_decomposition}) using the points in $\hat{G}^{l}_{n}$. Next, plane coefficient ${d}^{l}_{n}$ is calculated as ${d}^{l}_{n}=-(\mathbf{n}^{l}_{n})^{T}  {\bar{\mathbf{p}}^{l}_{n}}$\textcolor{black}{,} where $\bar{\mathbf{p}}^{l}_{n}$ denotes the mean point of $\hat{G}^{l}_{n}$. Finally, $\hat{G}^{l+1}_{n}$ is formulated as follows:

\vspace{-0.2cm}
\begin{equation}
\begin{aligned}
\hat{G}^{l+1}_{n}=\left\{\mathbf{p}_{k} \in S_{n} \mid {d}^{l}_{n} - \hat{{d}}_{k} < M_{d} \right.\}
\end{aligned}
\vspace{-0.1cm}
\end{equation}
where $\hat{d}_k=-(\mathbf{n}^{l}_{n})^{T} \mathbf{p}_k$ and $M_{d}$ denotes the distance margin of the plane. The procedure is repeated multiple times. According to Zermas \textit{et al.} \cite{zermas2017fast}, $\hat{G}_{n}=\hat{G}^{3}_{n}$ is empirically the final output on each $S_{n}$ in this paper.

\begin{figure}[t!]
	\centering
	\begin{subfigure}[b]{0.30\textwidth}
		\includegraphics[width=1.0\textwidth]{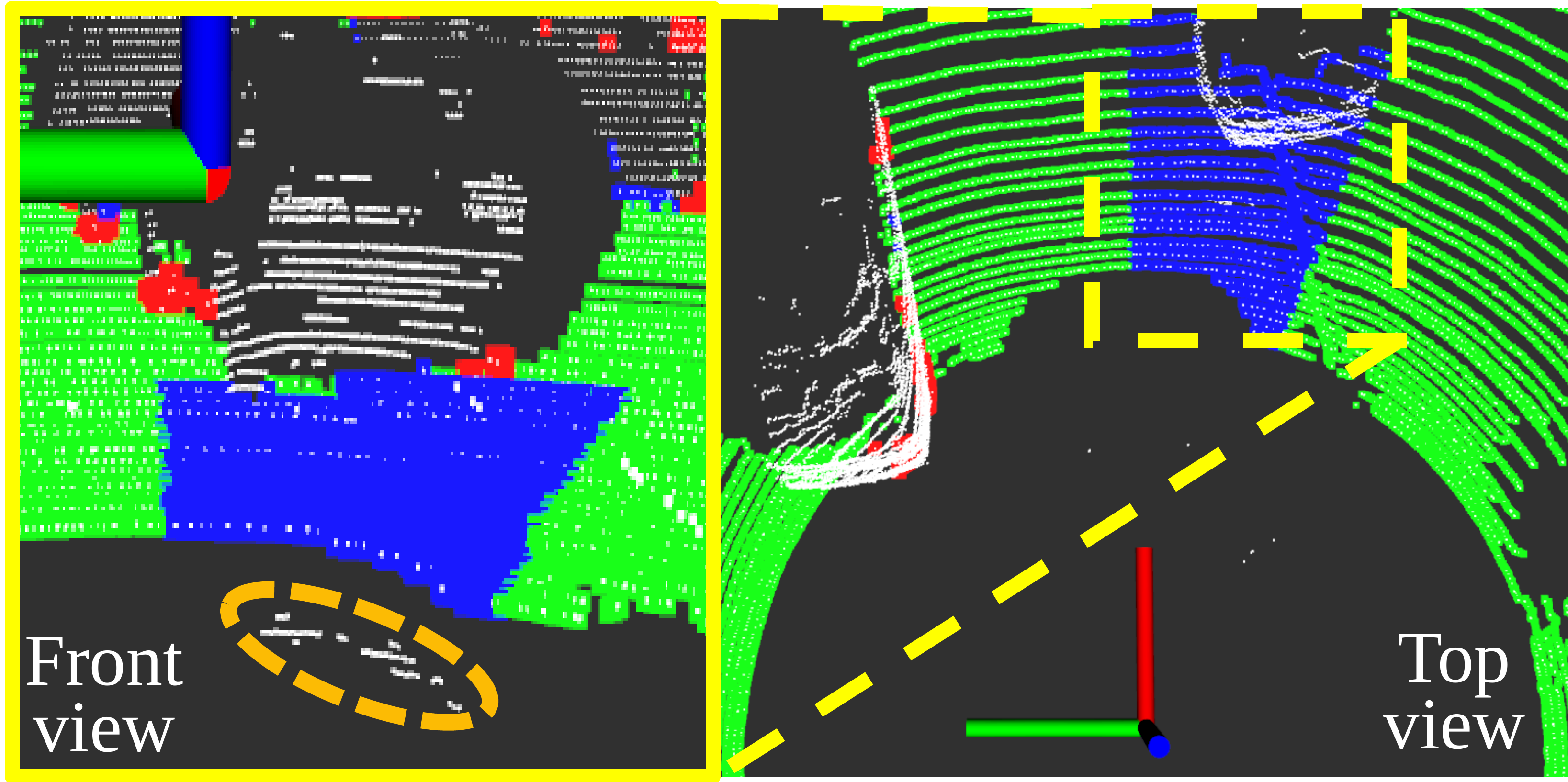}
		\caption{}
	\end{subfigure}
	\begin{subfigure}[b]{0.165\textwidth}
		\includegraphics[width=1.0\textwidth]{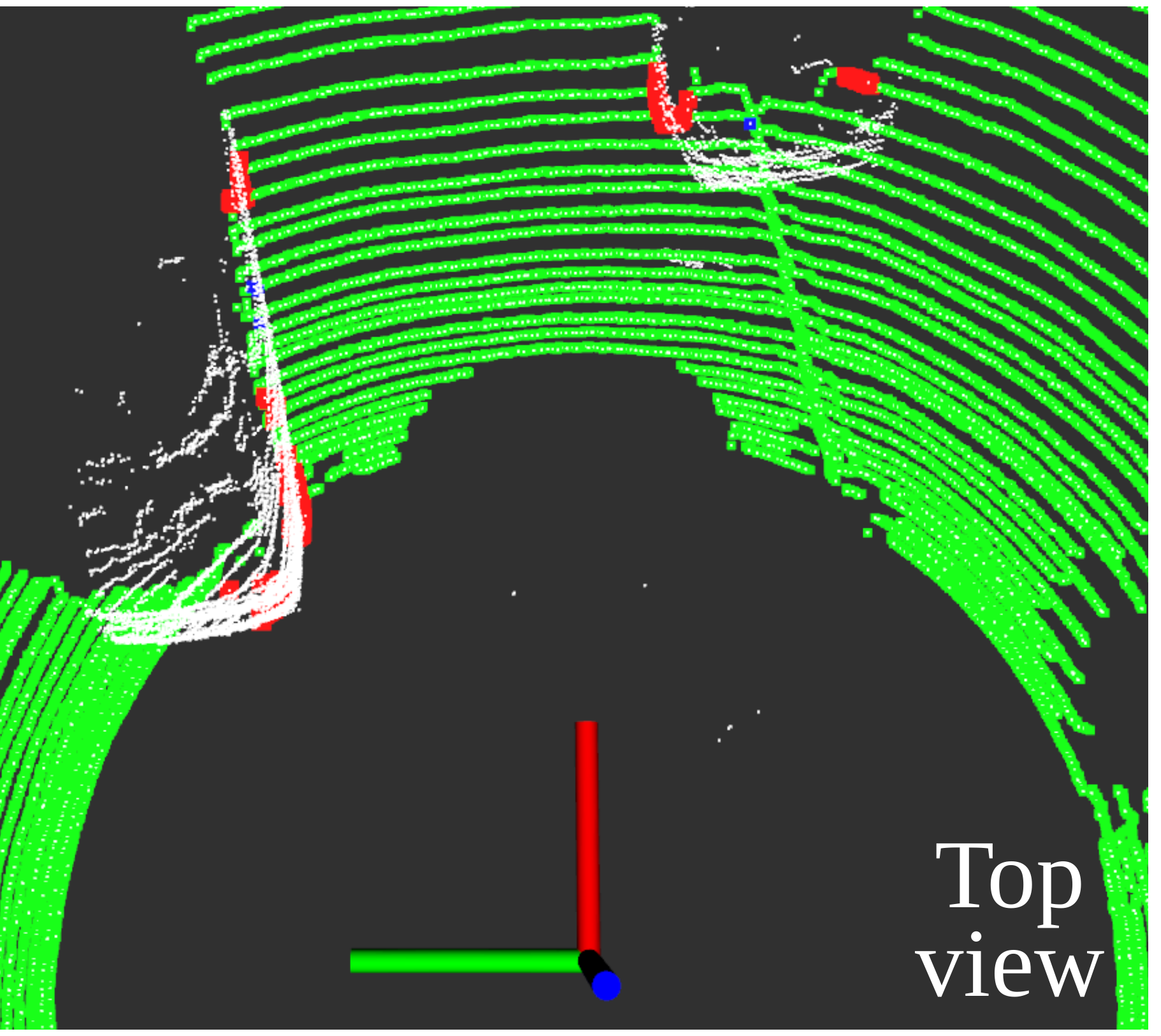}
		\caption{}
	\end{subfigure}
	\vspace{-0.25cm}
	\caption{(a) Before and (b) after the application of adaptive initial seed selection to prevent the effect of erroneous points, which are in dashed lines, on R-GPF for the SemanticKITTI dataset \cite{behley2019semantickitti} sequence \texttt{00} around frame 435. {In (a), the mismeasured points, which are inside the dashed circle, are sometimes selected as initial seeds and then cause the failure of region-wise ground plane fitting, which is represented by the blue region.} In this paper, the green, blue, and red points denote TPs, FNs, and FPs, respectively. {The less blue points are, the better it is} (best viewed in color).}
	\label{fig:adaptive_seed}
	\vspace{-0.3cm}
\end{figure}

Note that the main difference between the original R-GPF \cite{lim21erasor} and ours \textcolor{black}{is that ours involves using} adaptive initial seed selection to prevent R-GPF from converging to a local minimum. Occasionally, erroneous cloud points below the actual ground are acquired due to the multipath problem or reflection of LiDAR signals, as shown in Fig.~\ref{fig:adaptive_seed}(a) \cite{lim2019ronet}. It is observed that this phenomenon mostly happens in $Z_{1}$ because reflection occurs only in areas where the signal is relatively strong. These outliers impede R-GPF from estimating the right ground plane. 

 To address this issue, we utilize the fact that the {$z$} values of ground points only in $Z_1$ are mainly distributed near $-h_{\text{s}}$, where $h_{\text{s}}$ represents \textcolor{black}{the} sensor height. Accordingly, $\mathbf{p}_k$ in $S_n$ which belongs to $Z_1$ \textcolor{black}{is} filtered out if the $z_k$ is lower than $M_{h} \cdot h_{\text{s}}$ when \textcolor{black}{one is} estimating $\hat{G}^0_n$, where $M_{h}<-1$ is the height margin. For \textcolor{black}{an} $S_n$ that does not belong to $Z_1$, the adaptive threshold decreases as $m$ becomes larger to avoid \textcolor{black}{the} improper filtering of points that may come from downhill, which are actually TP. 



\vspace{-0.15cm}
\subsection{Ground Likelihood Estimation} 
\vspace{-0.1cm}
Consequently, it is necessary to robustly discern whether $\hat{G}_{n}$ belongs to the actual ground. To this end, \textcolor{black}{GLE} is proposed, which is a region-wise probabilistic test for binary classification. In doing so, Patchwork leverages GLE to improve \textcolor{black}{the} overall precision, excluding \textcolor{black}{the} initial unintended planes consisting of non-ground points.

Let $\mathcal{L}(\theta \mid \mathcal{X})$ be the GLE, where $\theta$ denotes all parameters of Patchwork and $\mathcal{X}$ represents a random variable following a continuous probability distribution with density function $f$. Let us presume that every bin is independent of each other. Then, $\mathcal{L}(\theta \mid \mathcal{X})$ is expressed as

\vspace{-0.3cm}
\begin{equation}
\mathcal{L}(\theta \mid \mathcal{X})=f(\mathcal{X} \mid \theta) = \prod_{n}  f(\mathcal{X}_{n} \mid \theta_{n})
\vspace{-0.15cm}
\end{equation}
where $\theta_{n}$ and $\mathcal{X}_{n}$ represent parameters and a random variable of each $\hat{G}_n$, respectively. Note that subscript $n$ denotes that the parameters are from $\hat{G}_n$.

Based on our prior knowledge, how likely each $\hat{G}_n$ is to be ground points is defined in terms of \textit{uprightness}, \textit{\textcolor{black}{elevation}}, and \textit{flatness}, which are represented as $\phi(\cdot)$, $\psi(\cdot)$, and $\varphi(\cdot)$, respectively, as follows: 

\vspace{-0.3cm}
\begin{equation}
\begin{split}
f(\mathcal{X}_n \mid \theta_n) \equiv  \phi({\mathbf{v}}_{3, n}) \cdot \psi(\bar{z}_n,r_n) \cdot \varphi(\psi(\bar{z}_n, r_n), \sigma_n) 
\end{split}
\vspace{-0.15cm}
\end{equation}
where $\bar{z}_n$, $r_n$, and $\sigma_n$ denote the mean {$z$} value, \textcolor{black}{the} distance between the origin and the centroid of a $S_{n}$, and the surface variable\textcolor{black}{,} where $\sigma_n = \frac{\lambda_{3,n}}{\lambda_{1,n} + \lambda_{2,n} + \lambda_{3,n}}$ \cite{eckart2018hgmr}, respectively. 

\begin{figure*}[t!]
	\centering
	\begin{subfigure}[b]{0.34\textwidth}
		\includegraphics[width=1.0\textwidth]{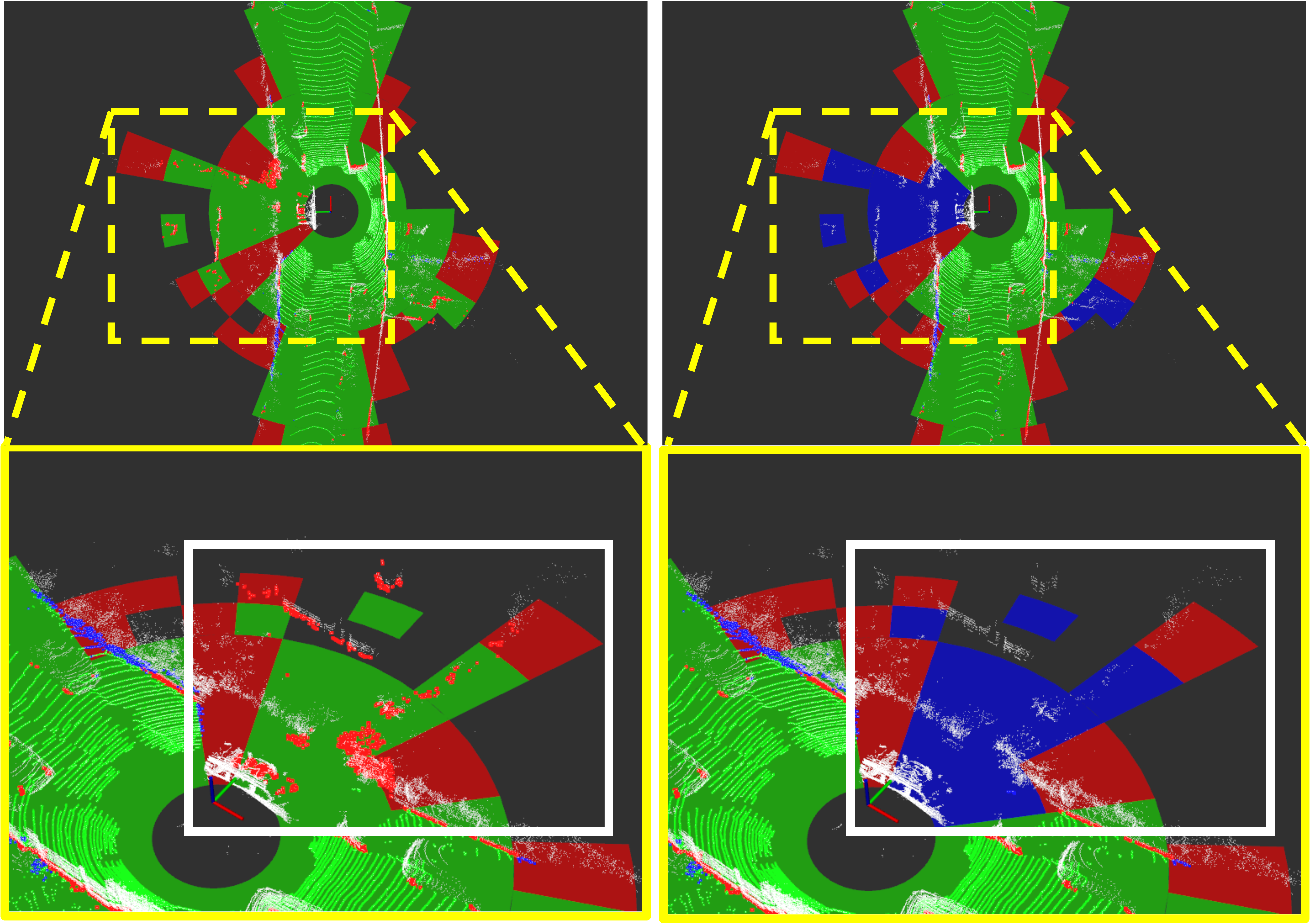}
		\caption{}
	\end{subfigure}
	\begin{subfigure}[b]{0.34\textwidth}
		\includegraphics[width=1.0\textwidth]{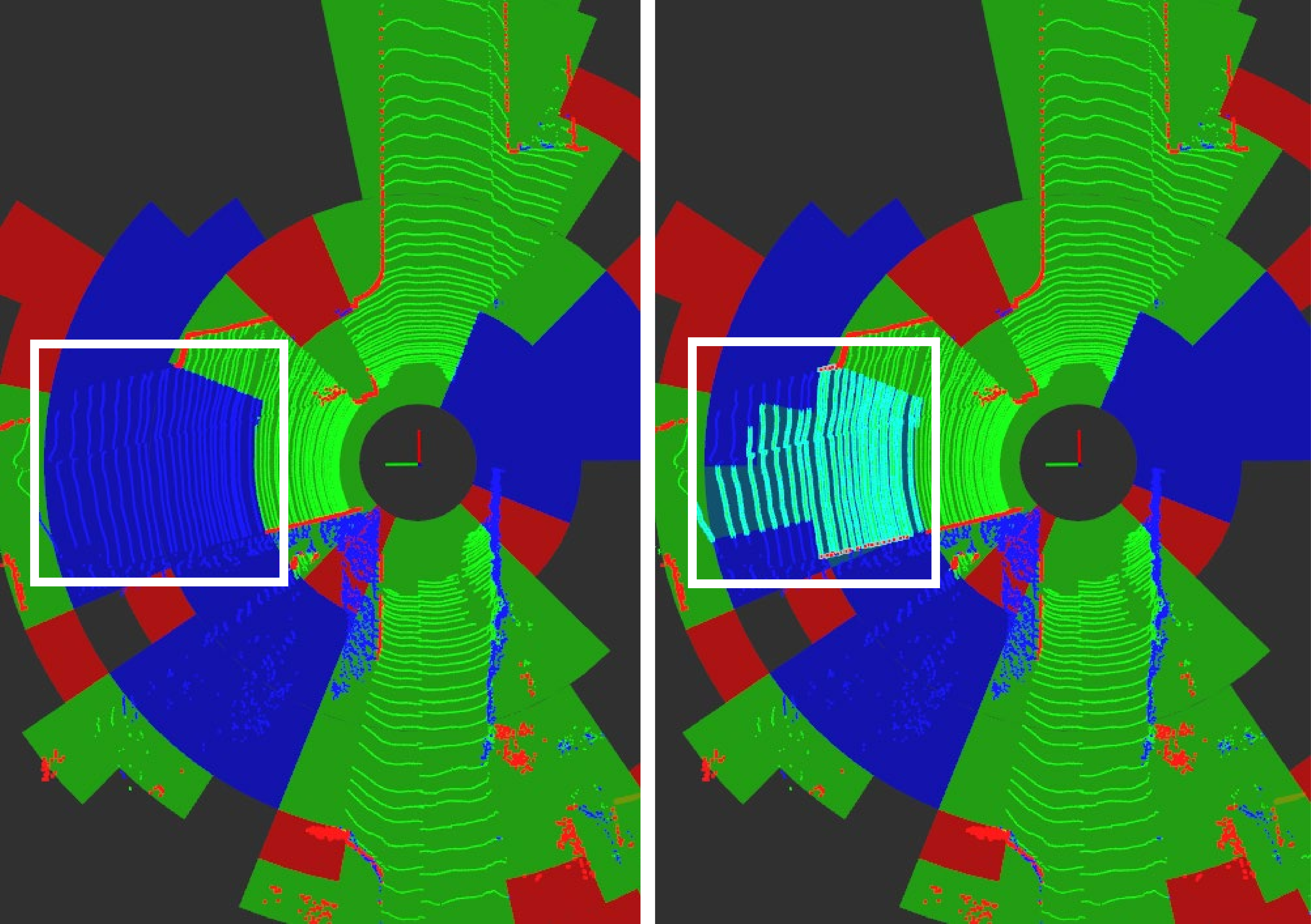}
		\caption{}
	\end{subfigure}
	\begin{subfigure}[b]{0.30\textwidth}
		\includegraphics[width=1.0\textwidth]{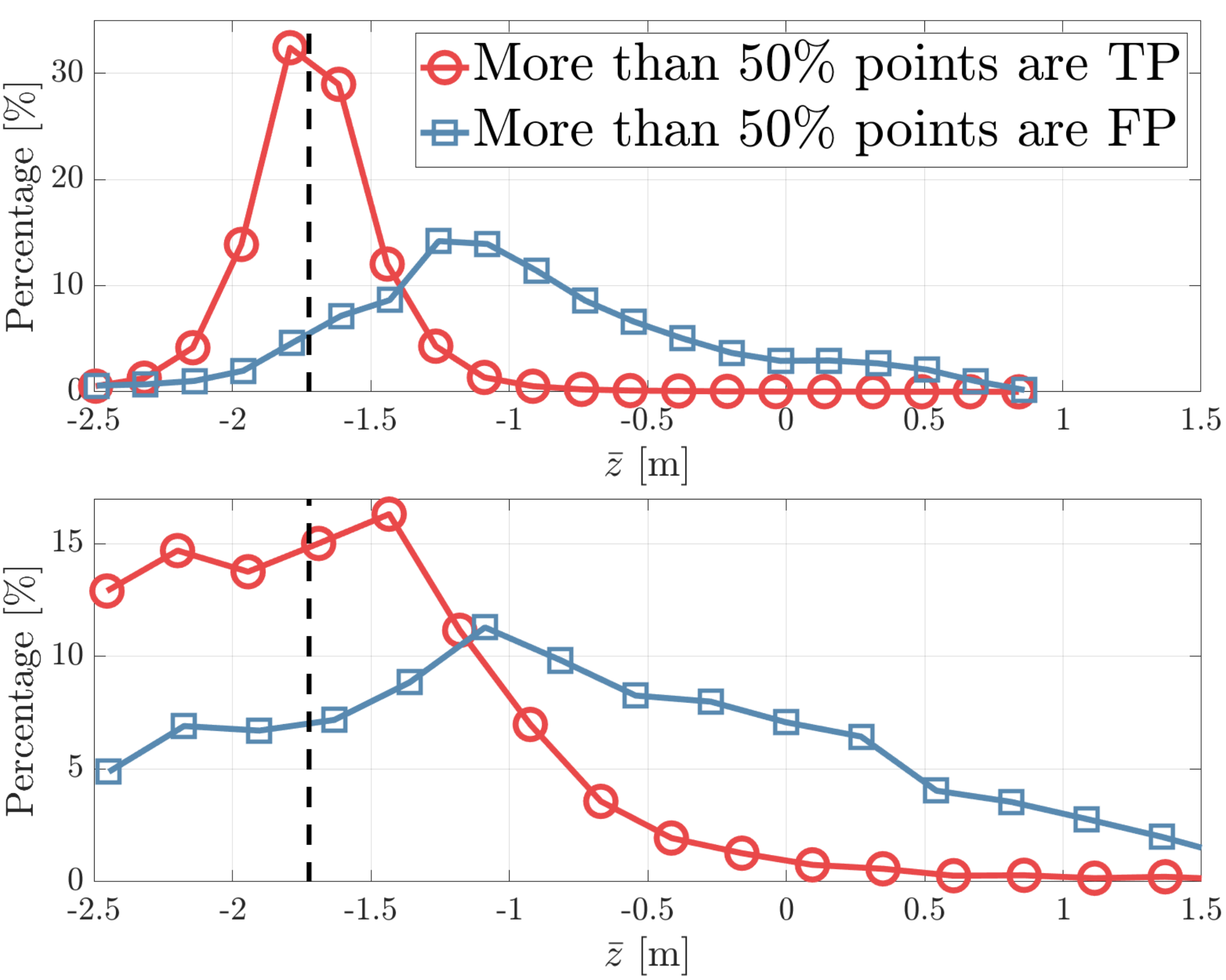}
		\caption{}
	\end{subfigure}
	
	\vspace{-0.2cm}
	\caption{(a) (L-R): Before and after the application of the \textcolor{black}{elevation filter} for sequence \texttt{00} around frame 2,810  on the SemanticKITTI dataset. Note that red cloud points which denote FPs are filtered. (b) (L-R): Before and after the application of flatness for sequence \texttt{10} around frame 286\textcolor{black}{,} where the cyan points denote reverted TPs via flatness, which are previously filtered through the \textcolor{black}{elevation}. The green, blue, and red regions denote regions where GLE is met, filtered by \textcolor{black}{the elevation}, and filtered by flatness, respectively. (c) (T-B): Probability Distribution Function (PDF) of the mean {$z$} value between two corresponding partial ground estimation\textcolor{black}{s} $\hat{G}_n$ by using uprightness alone in central zone $Z_1$ and quater zone $Z_2$, and outer zone $Z_4$ \textcolor{black}{on whole sequences of the SemanticKITTI dataset}. The dashed line represents \textcolor{black}{the} ground elevation from a sensor (best viewed in color).}
	\label{fig:GLE}
	\vspace{-0.4cm}
\end{figure*}

\noindent 
\textbf{Uprightness} \quad  Indeed, if $\hat{G}_n$ belongs to the actual ground (i.e. most of the points are in TP), it is observed that ${\mathbf{v}}_{3, n}$ is likely to be orthogonal to the ground \textcolor{black}{with} which terrestrial vehicles are in contact. In other words, ${\mathbf{v}}_{3, n}$ tends to be upright against the X-Y plane of the sensor frame. Therefore, \textcolor{black}{uprightness} indicator function is proposed to leverage the geometric characteristics as

\vspace{-0.2cm}
\begin{equation}
\phi({\mathbf{v}}_{3, n})=\left\{\begin{array}{ll}
1, &  \text{if} \; \frac{{\mathbf{v}}_{3, n} \cdot {\mathbf{z}}}{\norm{{\mathbf{v}}_{3, n}} \norm{{\mathbf{z}}}} > \cos(\frac{\pi}{2}-\theta_{\tau}) \\
0, &  \text{otherwise}
\end{array}\right.
\vspace{-0.15cm}
\end{equation}
where ${\mathbf{z}}=[0 \; 0 \;1]$ and $\theta_{\tau}$ is the uprightness margin\textcolor{black}{,} which denotes the angle between ${\mathbf{v}}_{3, n}$ and the X-Y plane. That is, the larger $\theta_{\tau}$ is, the more conservative the criterion becomes. As shown in Fig.~\ref{fig:GLE}(a) and (b), the red regions represent the cases where the uprightness has not been met, so $\phi({\mathbf{v}}_{3, n})$ is equal to zero. Through experiments, we set $\theta_{\tau}$ to $45^{\circ}$, which is empirically determined to be strict enough (see Section~\rom{4}.\textit{B}).

\noindent 
\textbf{\textcolor{black}{Elevation}} \quad  Unfortunately, it is not possible to filter \textcolor{black}{the} $\hat{G}_n$ belonging to the bonnet or roof of a car by using uprightness alone. \textcolor{black}{In addition}, when large objects, such as cars, are close to the sensor frame, occlusion occurs, thus giving rise to the partial observation issue. That is, partially measured cloud points above the occluded spaces are predicted as the $\hat{G}_n$, which are not\textcolor{black}{,} in fact\textcolor{black}{,} the lowest parts in the space. \textcolor{black}{This phenomenon is} presented in the left in Fig.~\ref{fig:GLE}(a).

To tackle this problem, a conditional logistic function, or $\psi(\bar{z}_n, r_n)$, is proposed. The key idea of the \textcolor{black}{elevation filter} is motivated by Asvadi \textit{et al.} \cite{asvadi2015detection_roof}\textcolor{black}{:} once $\bar{z}_n$ near the sensor frame is quite high compared with $-h_s$, $\hat{G}_n$ is likely not the ground. The experimental evidence supports our rationale as shown in Fig.~\ref{fig:GLE}(c). \textcolor{black}{U}sing uprightness alone, TPs and FPs become distinguishable based on $\bar{z}_n$ with a minor loss of TPs when $r_n$ is small, i.e. the case where $\hat{G}_n$ is in ${Z}_1$ and $Z_2$. In contrast, TPs and FPs are indistinguishable when $r_n$ is large, i.e. the case where $\hat{G}_n$ is in $Z_4$.

Based on these observations, $\psi(\bar{z}_n, r_n)$ is defined as follows:


\vspace{-0.5cm}
\begin{equation}
\psi(\bar{z}_n, r_n)=\left\{\begin{array}{ll}
(1+e^{(\bar{z}_n-\kappa (r_n) )})^{-1}, &  \text{if} \; r_n < L_{\tau} \\
1, &  \text{otherwise}
\end{array}\right.
\vspace{-0.2cm}
\end{equation}
where $\kappa (\cdot)$ denotes an adaptive midpoint function that exponentially increases depending on $r_n$. As visualized in Fig.~\ref{fig:GLE}(a), if $\bar{z}_n$ is lower than $\kappa(r)$, then the value of $\psi(\bar{z}_n, r_n)$ is \textcolor{black}{higher than} 0.5 when $r_n$ is smaller than constant range parameter $L_\tau$. Note that $\psi(\bar{z}_n, r_n)$ always becomes 1 when $r_n$ exceeds $L_\tau$ because it is \textcolor{black}{unclear} whether $\hat{G}_n$ comes from non-ground objects or \textcolor{black}{from} a steep incline as $r_n$ becomes larger.

\noindent 
\textbf{Flatness} \quad Finally, the aim of flatness is to revert some FNs filtered by the \textcolor{black}{elevation} if they are definitely an even plane. For instance, if $\hat{G}_n$ belongs to a very steep uphill, \textcolor{black}{and thus if} $\bar{z}_n$ is larger than $\kappa(r)$, \textcolor{black}{$\hat{G}_n$} is sometimes filtered out via $\psi(\bar{z}_n, r_n)$. To resolve this problem, we leverage the surface variable $\sigma_n$ to check \textcolor{black}{the} flatness of $\hat{G}_n$ which is considered to be FNs, even though $\psi(\bar{z}_n, r_n)$ is lower than 0.5. For this, \textcolor{black}{one possible implementation of} $\varphi(\psi(\bar{z}_n, r_n), \sigma_n)$ is defined as 


\vspace{-0.35cm}

\begin{equation}
\varphi(\psi(\bar{z}_n, r_n), \sigma_n)=\left\{\begin{array}{ll}
\zeta e^{-(\sigma_n - \sigma_{\tau,m})}, &  \text{if} \; \psi(\bar{z}_n, r_n) < 0.5 \\
1, &  \text{otherwise}
\end{array}\right.
\vspace{-0.15cm}
\end{equation}
where $\zeta > 1$ and $\sigma_{\tau,m}$ denote \textcolor{black}{the} magnitude of gain and the threshold of \textcolor{black}{the} surface variable depending on $Z_m$, respectively. \textcolor{black}{By doing so, the GLE of steep uphills increases, and they can be reverted into the ground estimation although $\bar{z}_n$ is higher than $\kappa(r_n)$.}

\begin{figure}[t!]
	\centering 
	\begin{subfigure}[b]{0.48\textwidth}
		\includegraphics[width=1.0\textwidth]{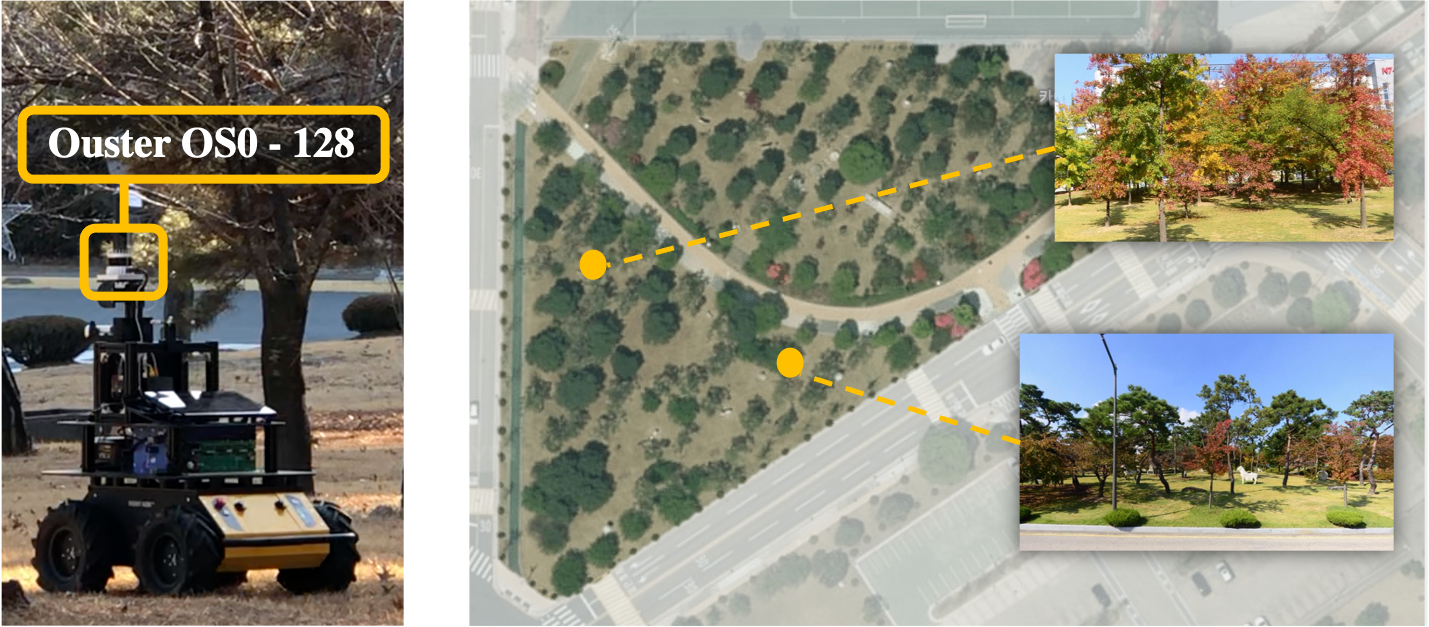}
	\end{subfigure}
	\caption{(L-R) Our robot platform for additional experiments. Rough terrain environment \textcolor{black}{on} KAIST campus.}
	\vspace{-0.3cm}
	\label{fig:our_nonplanar_experiment}
\end{figure}

Therefore, the final estimated ground points can be directly expressed as follows:

\vspace{-0.3cm}
\begin{equation}
\hat{G}=\bigcup_{n\in\left<N_{\mathcal{C}}\right>} \left[ f(\mathcal{X}_n \mid \theta_n) > 0.5 \right] \hat{G}_{n}
\vspace{-0.2cm}
\end{equation}
where $\left[ \cdot \right]$ denotes the Iverson bracket\textcolor{black}{,} which returns \textbf{true} if the condition is satisfied and \textbf{false} otherwise.

\section{Experiments}


\subsection{Dataset} \label{sec:exp_dataset}

\noindent \textbf{SemanticKITTI Dataset} \quad To evaluate the ground segmentation performance of our proposed method against other ground segmentation algorithms, we experimented \textcolor{black}{with} the SemanticKITTI dataset\cite{behley2019semantickitti}. Accordingly, the points annotated with selected classes, i.e. \texttt{lane marking}, \texttt{road}, \texttt{parking}, \texttt{sidewalk}, \texttt{other ground}, \texttt{vegetation}, and \texttt{terrain}, are considered \textcolor{black}{to be} ground-truth ground points to be extracted. Note that the \texttt{vegetation} contains leaves of trees, so only points whose {$z$} value\textcolor{black}{s} with respect to the sensor frame \textcolor{black}{are} below $-1.3$ m are considered \textcolor{black}{to be} ground truth\textcolor{black}{s}.

\noindent \textbf{Rough Terrain Dataset} \quad Even though the SemanticKITTI dataset represents various urban environments, data \textcolor{black}{were} acquired from vehicle platforms on pavements only. Therefore, we conducted an additional, more challenging experiment to prove the robustness and generality of our proposed algorithm. As shown in Fig.~\ref{fig:our_nonplanar_experiment}, our robot platform is equipped with a 3D LiDAR (Ouster OS0-128). The data \textcolor{black}{were} acquired from a rough terrain environment on the KAIST campus, Daejeon, South Korea.

\subsection{Error Metrics} 
To evaluate our proposed method quantitatively, \textit{Precision}, \textit{Recall}, \textit{$F_{1} \; score$}, and \textit{Accuracy} are considered. Let $N_{\text{TP}}$, $N_{\text{TN}}$, $N_{\text{FP}}$, and $N_{\text{FN}}$ be the number of points in TP, TN, FP, and FN, \textcolor{black}{respectively}; then, these metrics are defined as follows:
\begin{itemize}
\item Precision: $\frac{N_{\text{TP}}}{N_{\text{TP}}+N_{\text{FP}}}$, $\text{F}_1$ score: $\frac{2 \cdot  N_{\text{TP}}}{2 \cdot N_{\text{TP}}+N_{\text{FP}}+N_{\text{FN}}}$

\item Recall: $\frac{N_{\text{TP}}}{N_{\text{TP}}+N_{\text{FN}}}$, Accuracy: $\frac{N_{\text{TP}}+N_{\text{TN}}}{N_{\text{TP}}+N_{\text{TN}}+N_{\text{FP}}+N_{\text{FN}}}$

\end{itemize}

\subsection{Parameters of Patchwork} 

We set $\{N_{r, 1},N_{r, 2},N_{r, 3},N_{r, 4}\}=\{2,4,4,4\}$, $\{N_{\theta, 1}, N_{\theta, 2}, N_{r\theta, 3}, N_{\theta, 4}\}=\{16, 32, 54, 32\}$, $L_{\min}=2.7\text{m}$, and $L_{\max}= 80.0\text{m}$ for the \textcolor{black}{CZM}. For R-GPF, $N_{\text{seed}}=20$, $z_{\text{seed}}=0.5$, $M_{d}=0.15$ and $M_{h}=-1.1$.  The smaller $z_{\text{seed}}$ and $M_{d}$ are, the more conservative the criterion becomes. Finally, for GLE, $L_\tau=L_{\max, 2}$, $\sigma_{\tau,1}=0.00012$, and $\sigma_{\tau,2}=0.0002$, for which cloud points whose $\sigma_{\tau}$ is lower than 0.01 are considered \textcolor{black}{to be} a plane \cite{eckart2018hgmr}, yet we set more strict criteria.

\section{Results and Discussion} \label{sec:results}

\subsection{Performance Analysis with Different Bin Sizes}

\begin{figure}[t!]
	\centering 
	\begin{subfigure}[b]{0.50\textwidth}
		\includegraphics[width=1.0\textwidth]{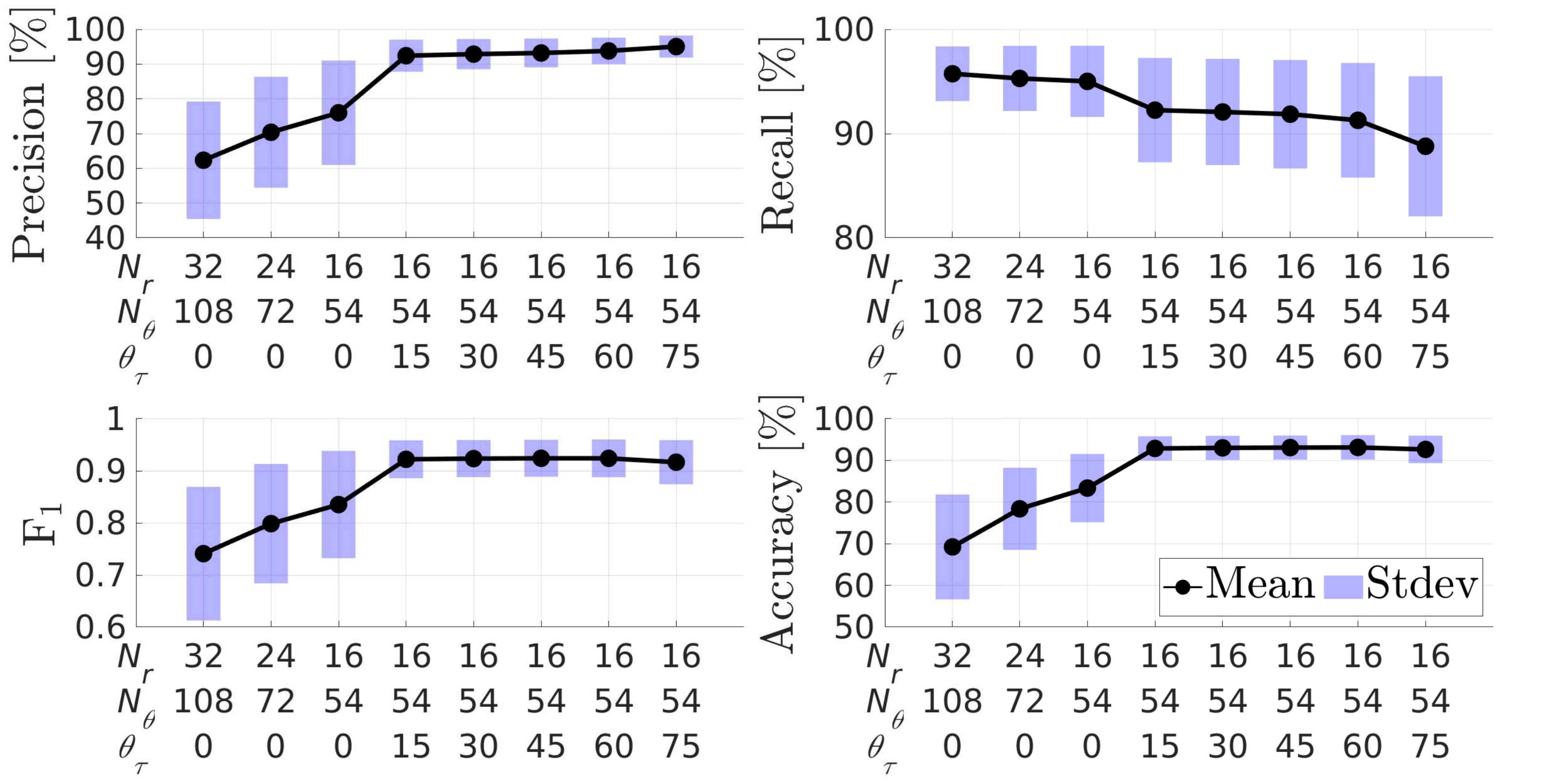}
	\end{subfigure}
	\vspace{-0.1cm}
	\caption{Performance changes of uniform polar representation with varying bin sizes $(N_{r}, N_{\theta})$ and uprightness threshold $\theta_{\tau}$ on the SemanticKITTI dataset.}
	\label{fig:parameter_analysis}
	\vspace{-0.5cm}
\end{figure}

First, the effect of bin size is analyzed, which was not conducted in our previous study \cite{lim21erasor}. As shown in Fig.~\ref{fig:parameter_analysis}, a larger bin size leads to a large improvement \textcolor{black}{in} precision with little \textcolor{black}{recall} degradation; thus, it raises the $\text{F}_{1}$ score dramatically. This result implies that the larger the \textcolor{black}{bin} area \textcolor{black}{is}, the more general\textcolor{black}{ly the} ground can be estimated. However, it \textcolor{black}{leads to} lower recall \textcolor{black}{as the bin size becomes larger} due to the fact that the resolution of each bin is decreased.

\subsection{Impact of Uprightness}

Unfortunately, \textcolor{black}{the} na\"ive enlargement of bin size does not resolve the large variance of precision, as presented in Fig.~\ref{fig:parameter_analysis}. However, using only uprightness leads to \textcolor{black}{a} dramatic performance improvement in precision while reducing the variance and thus improves the $\text{F}_{1}$ score. Thus, the result shows that our uprightness successfully filters \textcolor{black}{out the} wrongly estimated partial ground. However, \textcolor{black}{a} higher $\theta_{\tau}$ allows actual TPs, such as steep slope regions or boundary regions between road\textcolor{black}{s} and curbs, to be classified as non-ground, and thus recall is slightly decreased. Therefore, we can conclude that the $\theta_{\tau}$ of $45^\circ$ yielded the most reasonable estimation throughout the experiment.


\subsection{Effectiveness of Ground Likelihood Estimation}

\begin{figure}[t!]
	\centering 
	\begin{subfigure}[b]{0.49\textwidth}
		\includegraphics[width=1.0\textwidth,trim={0.2cm 0.2cm 0 0},clip]{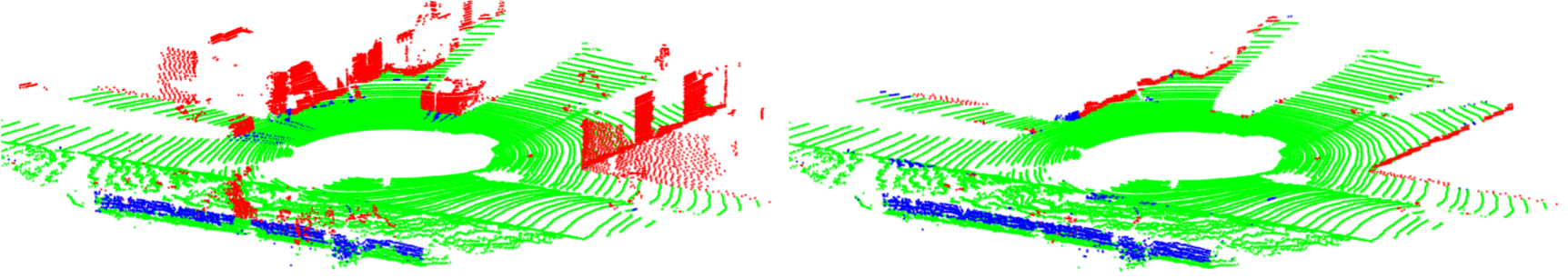}
	\end{subfigure}
	\caption{(L) Ground estimation results from R-GPF \cite{lim21erasor} and (R) Patchwork for sequence \texttt{00} around frame 429 on the SemanticKITTI dataset. Our proposed GLE successfully rejects FPs, \textcolor{black}{with} green, blue, and red points denoting TPs, FNs, and FPs, respectively. {The less red points are, the better it is} (best viewed in color).}
	\label{fig:rgpf_vs_patchwork}
	\vspace{-0.1cm}
\end{figure}

\begin{figure}[t!]
	\centering 
	\begin{subfigure}[b]{0.49\textwidth}
		\includegraphics[width=1.0\textwidth]{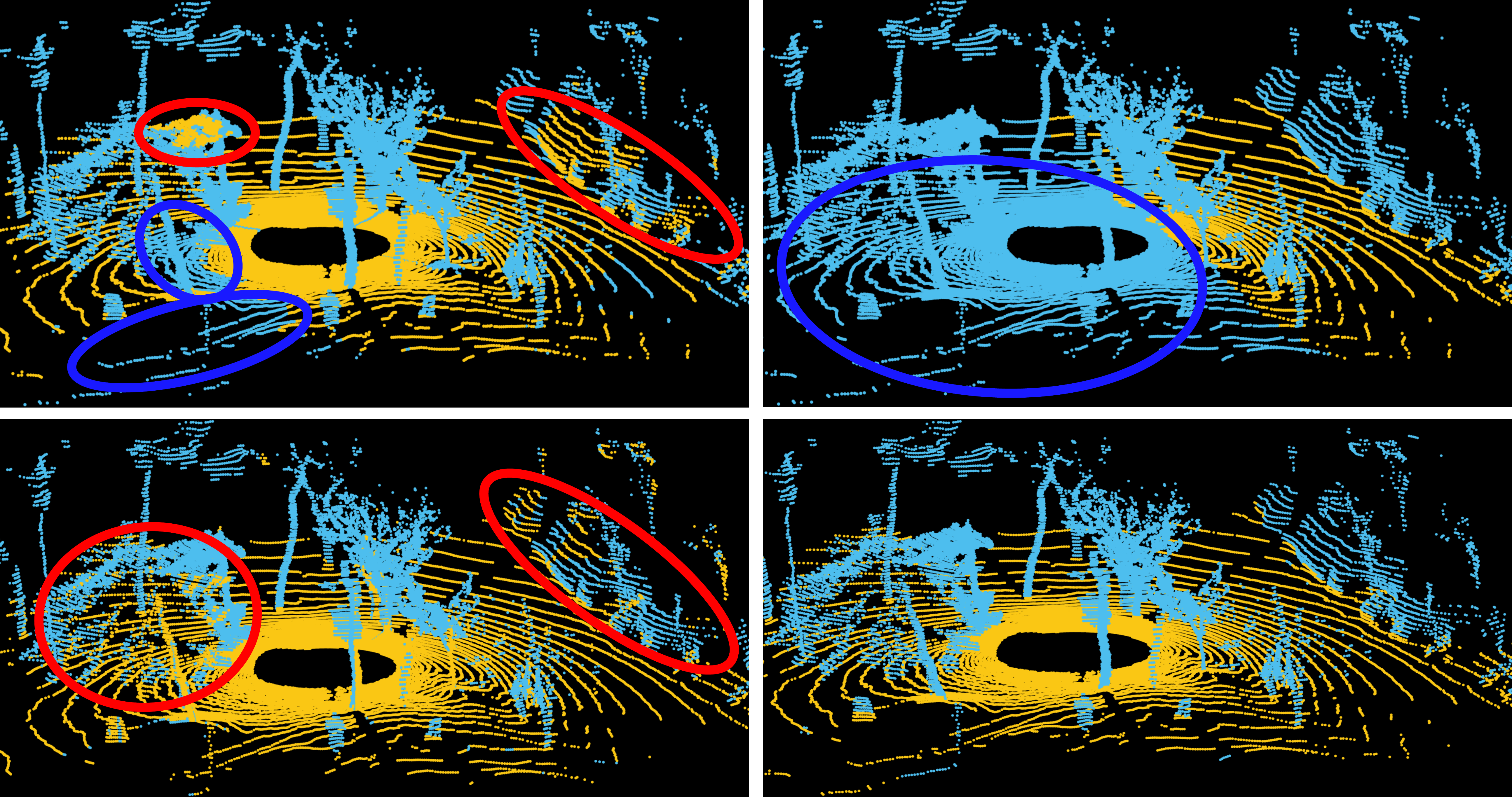}
	\end{subfigure}
	\vspace{-0.2cm}
	\caption{{Qualitative} comparison of multiple plane fitting--based methods. (L-R, T-B): Output from GPF~\cite{zermas2017fast}, CascadedSeg~\cite{narksri2018slope}, R-GPF~\cite{lim21erasor}, and Patchwork on rough terrain. Our method shows its robustness even though the ground is bumpy and sloped. {The yellow and cyan points denote estimated ground and non-ground, respectively. The points inside blue circles denote FNs and those inside red circles denote FPs} (best viewed in color).}
	\label{fig:lig_quantatitve_comparison}
	\vspace{-0.1cm}
\end{figure}

\begin{figure*}[t!]
	\centering
	\begin{subfigure}[b]{0.16\textwidth}
		\includegraphics[width=1.0\textwidth]{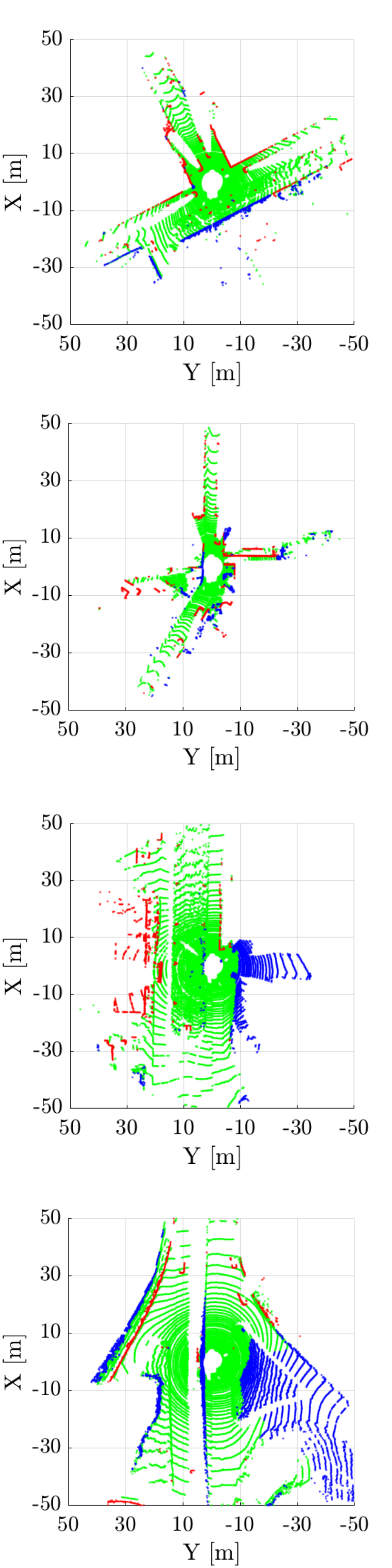}
		\caption{RANSAC \cite{fischler1981ransac}}
	\end{subfigure}
	\begin{subfigure}[b]{0.16\textwidth}
		\includegraphics[width=1.0\textwidth]{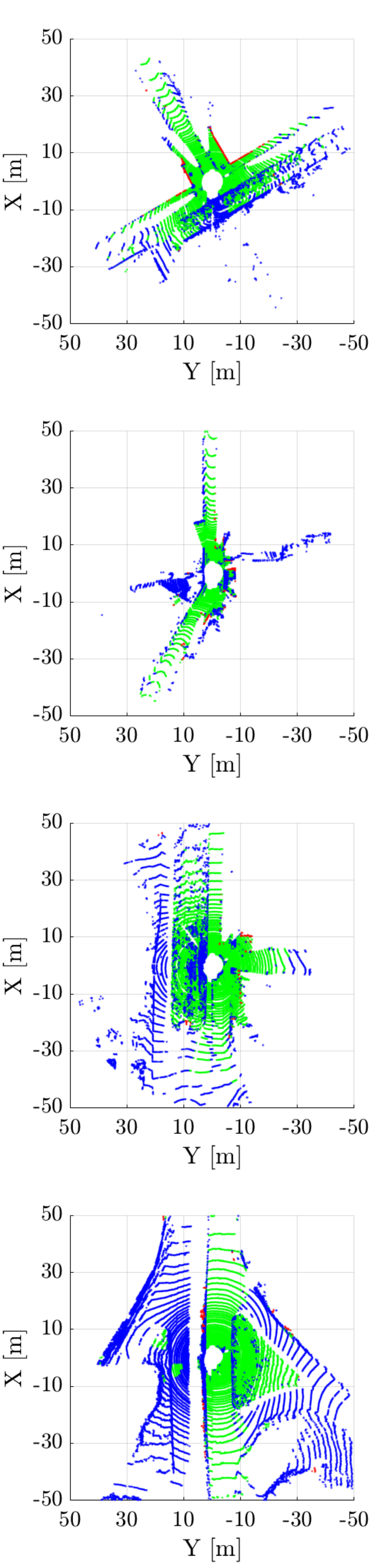}
		\caption{LineFit \cite{himmelsbach2010fast}}
	\end{subfigure}
	\begin{subfigure}[b]{0.16\textwidth}
		\includegraphics[width=1.0\textwidth]{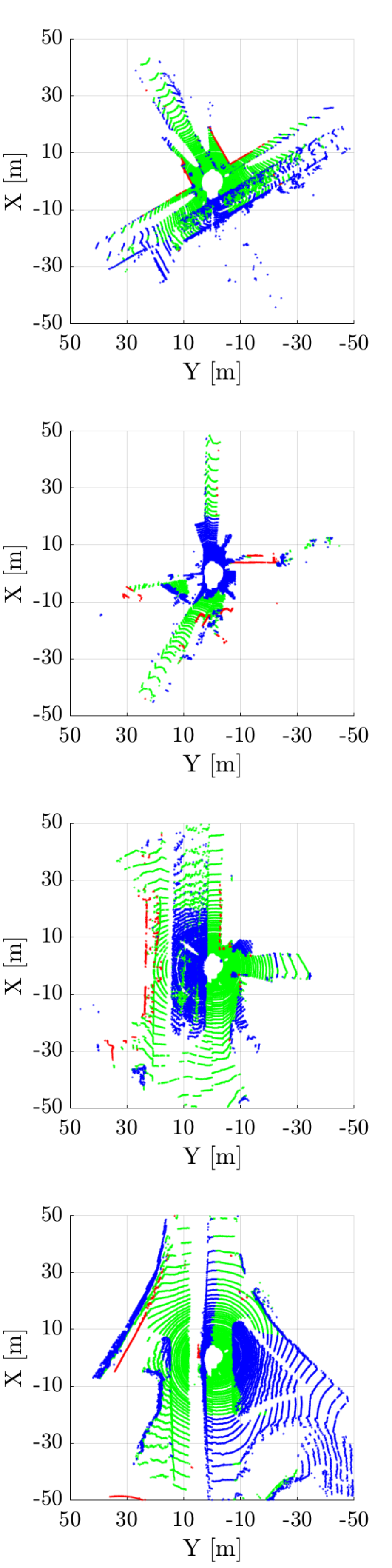}
		\caption{GPF \cite{zermas2017fast}}
	\end{subfigure}
	\begin{subfigure}[b]{0.16\textwidth}
		\includegraphics[width=1.0\textwidth]{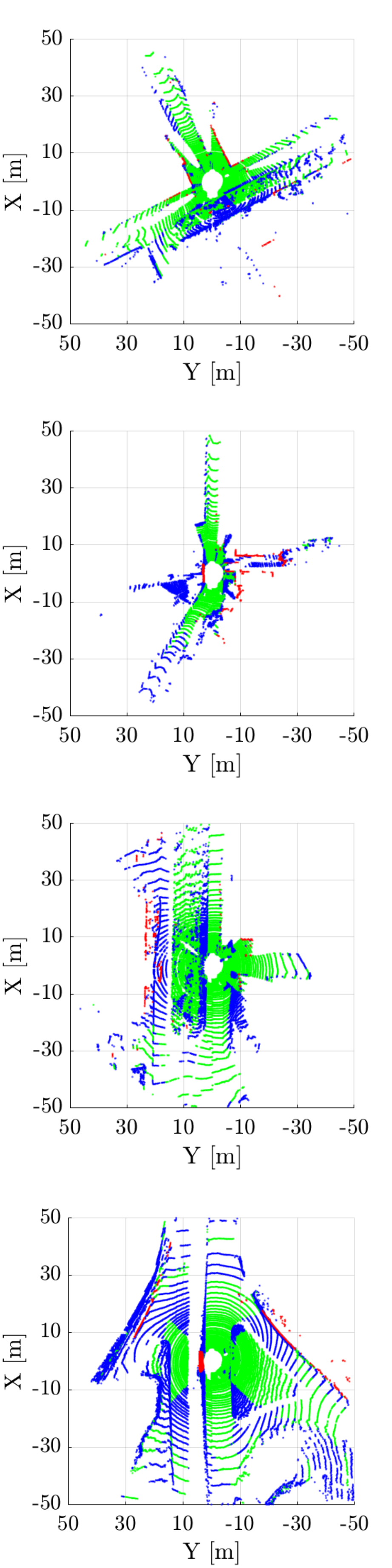}
		\caption{CascadedSeg \cite{narksri2018slope}}
	\end{subfigure}
	\begin{subfigure}[b]{0.16\textwidth}
		\includegraphics[width=1.0\textwidth]{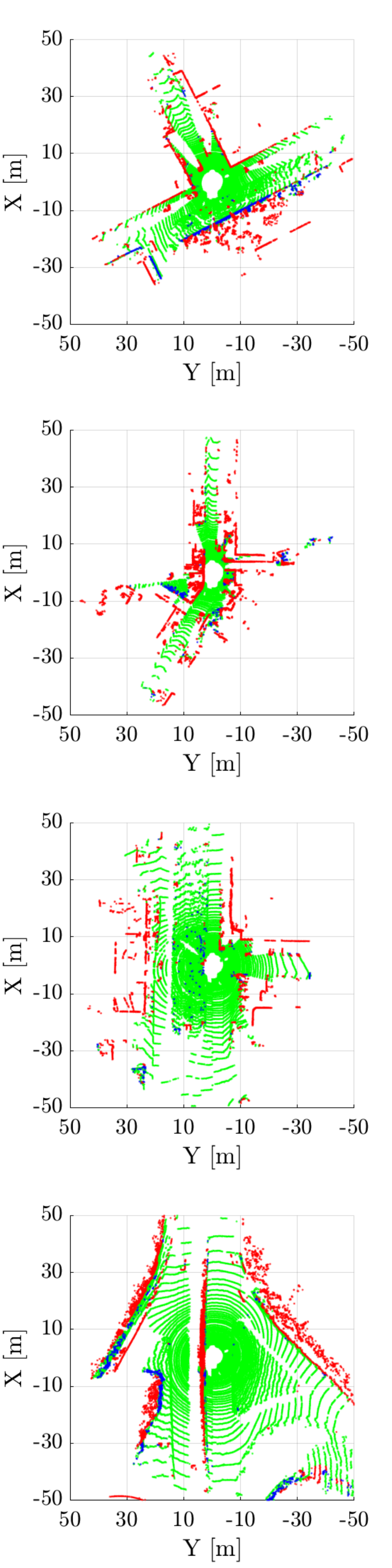}
		\caption{R-GPF \cite{lim21erasor}}
	\end{subfigure}
	\begin{subfigure}[b]{0.16\textwidth}
		\includegraphics[width=1.0\textwidth]{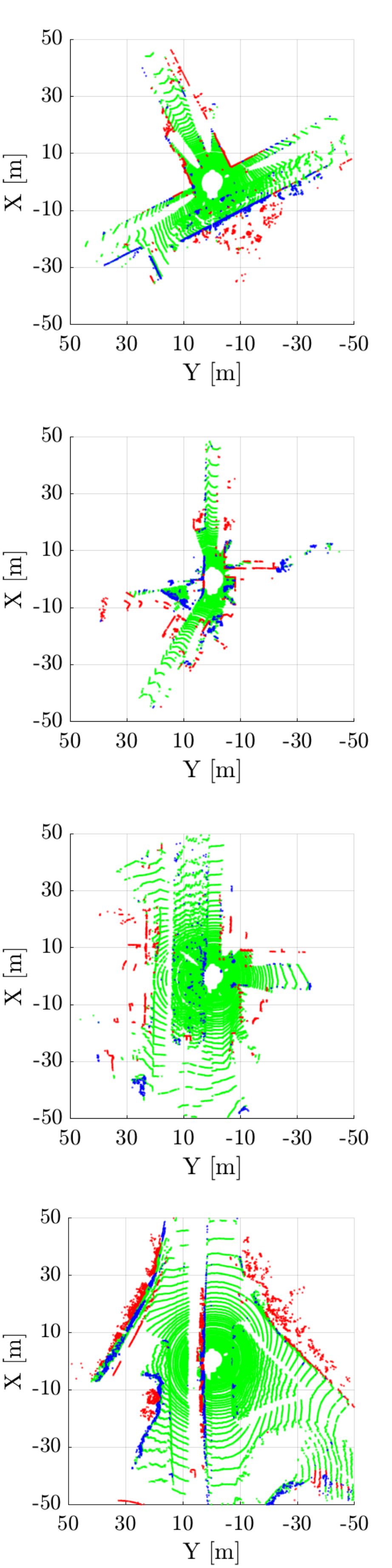}
		\caption{Proposed}
	\end{subfigure}
	\caption{Comparison of ground estimation results produced by the proposed method and state-of-the-arts methods on the Semantic KITTI dataset (T-B): sequence \texttt{00} around frame\textcolor{black}{s} 429 and 1,800, \texttt{06} around frame 505, and \texttt{01} around frame 180. The green, blue, and red points denote TPs, FNs, and FPs, respectively. {For blue and red points, the less, the better and for green points, the more, the better} (best viewed in color).}
	\label{fig:sota_comparison}
	\vspace{-0.2cm}
\end{figure*}

 Thereafter, the effectiveness of GLE is validated. As shown in Fig.~\ref{fig:rgpf_vs_patchwork}, Fig.~\ref{fig:lig_quantatitve_comparison}{, and Table~\ref{table:sota_comparison},} R-GPF, which is our baseline algorithm, estimates \textcolor{black}{a} ground plane with many FP\textcolor{black}{s} because R-GPF gives priority to \textcolor{black}{the} maximization of recall. \textcolor{black}{Meanwhile}, our GLE successfully filters out the wrongly estimated partial ground, \textcolor{black}{thus dramatically} reducing FN\textcolor{black}{s}. In particular, it is verified that many walls and a parked car can be rejected by the uprightness and \textcolor{black}{elevation filters}, respectively. 

\vspace{-0.2cm}
\subsection{Comparison with State-of-the-\textcolor{black}{A}rt Methods}
\vspace{-0.1cm}

Patchwork was quantitatively compared with state-of-the-art methods, namely, RANSAC \cite{fischler1981ransac}, LineFit\footnote{https://github.com/lorenwel/linefit\_ground\_segmentation} \cite{himmelsbach2010fast}, GPF\footnote{https://github.com/VincentCheungM/Run\_based\_segmentation} \cite{zermas2017fast}, and CascadedSeg\footnote{https://github.com/n-patiphon/cascaded\_ground\_seg} \cite{narksri2018slope}. We leveraged the open\textcolor{black}{--}source implementations for the experiment.

 \begin{table}[t!]
    \centering
    \caption{Performance comparison with the state-of-the-art methods on whole sequences of \textcolor{black}{the} SemanticKITTI dataset. {{\texttt{U}, \texttt{E}, and \texttt{F} denote the uprightness, elevation, and flatness terms, respectively; the red color denotes low Precision/Recall and large standard deviations.}}}
    \vspace{-0.1cm}
    \begin{tabular}{lccccc}
		\toprule
	\multirow{2}{*}{Method} & \multicolumn{2}{c}{Precision [\%]} & \multicolumn{2}{c}{Recall [\%]} & \multirow{2}{*}{$\text{F}_1$} \\  \cmidrule(lr){2-3} \cmidrule(lr){4-5}  
	&     $\mu$ & $\sigma$ & $\mu$ & $\sigma$ &  \\ \midrule
 	RANSAC \cite{fischler1981ransac}  & 88.16   & \textcolor{red}{17.18} &  91.26 & \textcolor{red}{12.52} & 0.897 \\
    GPF \cite{zermas2017fast}  & 95.84 & 3.41 &  \textcolor{red}{79.57} & \textcolor{red}{21.08} & 0.870 \\
    LineFit \cite{himmelsbach2010fast}  & \textbf{98.17} & \textbf{1.47} &  \textcolor{red}{78.26} & \textcolor{red}{9.29} & 0.871  \\
    CascadedSeg \cite{narksri2018slope} & 91.16 & \textcolor{red}{11.10} &  \textcolor{red}{68.95} & \textcolor{red}{11.18} & 0.785  \\ 
    R-GPF \cite{lim21erasor} & \textcolor{red}{66.94} & \textcolor{red}{16.50} &  93.34 & \textbf{3.41} & 0.780  \\ \midrule
    R-GPF \cite{lim21erasor} + \texttt{U} & 91.88 & 4.97 &  92.84 & 4.45 & 0.924  \\ 
    CZM + \texttt{U}  & 91.38 & 5.33 &  \textbf{93.65} & 4.10 & 0.925  \\ 
    CZM + \texttt{U} + \texttt{E}  & 92.89 & 4.25 &  93.08 & 4.73 & \textbf{0.930}  \\ 
    CZM + \texttt{U} + \texttt{E} + \texttt{F} & 92.47 & 4.26 &  93.43 & 4.63 & \textbf{0.930}  \\
    \bottomrule
	\end{tabular}
	\label{table:sota_comparison}
	\vspace{-0.4cm}
\end{table}
 
As shown in Fig.~\ref{fig:sota_comparison}, other methods show detailed ground estimation\textcolor{black}{s}. However, they struggle with non-planar regions, including a steep slope, a complex intersection, and a region where many curbs exist. In particular, LineFit is likely to be sensitive when encountering undulated terrains or bushy regions \cite{narksri2018slope}, estimating many FNs. On the other hand, some bins in GPF and CascadeSeg sometimes tend to converge to a local minimum because the bin size is too big, so it is not safe to assume that the ground is planar within the bin; in particular, this phenomenon becomes worse in rough terrain as presented in Fig.~\ref{fig:lig_quantatitve_comparison}. Thus, they show the large variance of recall on the SemanticKITTI dataset as shown in Table~\ref{table:sota_comparison}.

 In contrast, our proposed method shows promising performance. \textcolor{black}{In particular}, our method estimates the ground with little variance of recall relative to other methods. This confirms that our method overcomes the under-segmentation problem and is thus robust against these corner cases in urban environments. 
 
 Meanwhile, one may observe from Fig.~\ref{fig:sota_comparison} that the number of FN\textcolor{black}{s} from Patchwork is larger than th\textcolor{black}{ose} of LinFit, GPF, and CascadedSeg, yet these are actually the lowest parts of some objects, as shown in Fig.~\ref{fig:rgpf_vs_patchwork}. Th\textcolor{black}{is} means \textcolor{black}{that} even though FN degrades the performance of quantitative metrics, they rather help \textcolor{black}{to} resolve the under-segmentation issue.








\subsection{Algorithm speed} \label{sec:korea_univ}

To check the speed of each algorithm, \textcolor{black}{we used Intel(R) Core(TM) i7-7700K CPU}. Note that our proposed method shows the fastest speed among various multiple plane fitting--based methods as shown in Table~\ref{table:kitti_speed}. In particular, it is surprising that Patchwork is faster than R-GPF. This is because our \textcolor{black}{CZM} reduces the number of bins, so the amount of plane fitting is also decreased; for instance, Patchwork uses 504 bins, \textcolor{black}{whereas} R-GPF uses 3,240 bins. Furthermore, our method is based on PCA, so it is also faster than CascadedSeg\textcolor{black}{,} which is a RANSAC--based method. Therefore, this result shows that our method is not only robust but also fast enough for preprocessing use.

\begin{table}[t!]
	\centering
	\caption{Mean algorithm speed of multiple grid-based methods on sequence \texttt{05} of the SemanticKITTI dataset.}
	\vspace{-0.2cm}
	\begin{tabular}{lclc}
	\toprule
	Category & Baseline &Method & Runtime [Hz]   \\ \midrule
	Line fitting & - & LineFit \cite{himmelsbach2010fast} & \textbf{58.96} \\ \midrule
	\multirow{4}{*}{Plane fitting} & RANSAC & CascadedSeg \cite{narksri2018slope} & 13.07 \\ \cmidrule{2-4}
	&\multirow{3}{*}{PCA} & GPF \cite{zermas2017fast} & 29.72 \\ 
	& & R-GPF \cite{lim21erasor} & 35.30  \\
	& & Patchwork (Ours) & \textbf{43.97}  \\  \bottomrule
	\end{tabular}
	\label{table:kitti_speed}
	\vspace{-0.4cm}
\end{table}

\section{Conclusion}
\vspace{-0.1cm}
In this study, a fast and robust ground segmentation method, Patchwork, has been proposed. Our proposed method was proved to overcome the under-segmentation problem compared with previous approaches. In particular, our method provides a well-segmented ground estimation with smaller variations in performance, which enables mobile robots to detect non-ground objects in a robust way. In future works, we plan to apply our Patchwork to the detection of moving objects or to devise a deep learning--aided ground likelihood estimation for more sophisticated ground segmentation.

\bibliographystyle{IEEEtran}
\bibliography{./icra21,./IEEEabrv}

\end{document}